# Structure and Complexity in Planning with Unary Operators

**Ronen I. Brafman**                                   BRAFMAN@CS.BGU.AC.IL
**Carmel Domshlak**                                    DCARMEL@CS.BGU.AC.IL
*Department of Computer Science*
*Ben-Gurion University*
*P.O. Box 653, 84105 Beer-Sheva, Israel*

## Abstract

Unary operator domains – i.e., domains in which operators have a single effect – arise naturally in many control problems. In its most general form, the problem of STRIPS planning in unary operator domains is known to be as hard as the general STRIPS planning problem – both are PSPACE-complete. However, unary operator domains induce a natural structure, called the domain's *causal graph*. This graph relates between the preconditions and effect of each domain operator. Causal graphs were exploited by Williams and Nayak in order to analyze plan generation for one of the controllers in NASA's Deep-Space One spacecraft. There, they utilized the fact that when this graph is acyclic, a serialization ordering over any subgoal can be obtained quickly. In this paper we conduct a comprehensive study of the relationship between the structure of a domain's causal graph and the complexity of planning in this domain. On the positive side, we show that a non-trivial polynomial time plan generation algorithm exists for domains whose causal graph induces a polytree with a constant bound on its node indegree. On the negative side, we show that even plan existence is hard when the graph is a directed-path singly connected DAG. More generally, we show that the number of paths in the causal graph is closely related to the complexity of planning in the associated domain. Finally we relate our results to the question of complexity of planning with serializable subgoals.

## 1. Introduction

One of the first well formulated problems addressed by AI researchers was the planning problem. Simply stated, it involves the generation of a sequence of system transformations, taken out of a given set of system transformations (called *actions* or *plan operators*), whose combined effect is to move the system from some given initial state into one of a set of desired goal states. The planning problem is known to be intractable in general (Chapman, 1987), and tractable algorithms exist for very restrictive classes of problems only. This discouraging fact has not deterred planning researchers. Indeed, many researchers believe that real-world problems have some properties, or *structure*, that could be exploited, either implicitly or explicitly. In this paper we attempt to understand the relationship between structure and complexity in planning problems in which each action changes the value of a single variable.

To study the relation between the structure and the complexity in a class of problems we must identify a set of parameters that characterize it. In the case of planning, a number of such problem properties have been studied in the past (which we review in more detail in Section 6). These properties have been mostly syntactical, i.e., they involve restriction on operators, e.g., the type and number of preconditions or effects that operators have. For example, Bylander (1994) showed that STRIPS planning in domains where each operator is restricted to have positive preconditions and one postcondition only is tractable. Bäckström and Klein (1991b) considered other, more global types of syntactical restrictions, but using a more refined model in which two types of preconditions are considered: *prevail* conditions, which are variable values that are required prior to the execution of the operator and are not affected by the operator, and *preconditions*, which *are* affected by the operator. For example, they have shown that when operators have a single effect, no two operators have the same effect, and each variable can be affected only in one context (of prevail conditions) then the planning problem can be solved in polynomial time. However, these restrictions are very strict, and it is difficult to find reasonable domains satisfying them.

In this paper we concentrate on more global properties of unary operator domains; properties that capture some of the interactions between different planning operators. The tool we use to study these properties is the domains' *causal graph*. A causal graph is a directed graph whose nodes stand for the domain propositions. An edge $(p, q)$ appears in the causal graph if and only if some operator that changes the value of $q$ has a prevail condition involving $p$. Such a problem structure was introduced by Knoblock (1994) in the context of automatically generating abstractions for planning. Subsequently, Jonsson and Bäckström (1998b) introduced the 3S class of planning problems with unary operators, which was characterized by the acyclicity of the causal graph, and some restrictions on the operator set. It was shown that determining plan existence for this class of problems is polynomial, while plan generation is provably intractable.

Complexity results for unary operators would be of theoretical interest alone if one could not supply interesting problems in which unary operators are used. One interesting application in which this problem arises is the determination of dominance relationship between different outcomes in a CP-net (Boutilier, Brafman, Hoos, & Poole, 1999). This problem is reducible to STRIPS planning with unary operators.

Another example, of greater interest to the planning community, is a planning-based reactive control system that commands the NASA Deep Space One autonomous space-craft (Pell, Bernard, Chien, Gat, Muscettola, Nayak, Wagner, & Williams, 1997; Williams & Nayak, 1996, 1997). This system was hailed by Weld (1999) in his recent survey of AI planning as one of the most exciting recent developments in the area of planning. Naturally, the complete system (Pell et al., 1997) is very complex, however, its configuration planning and execution subsystem are of particular interest to us. In the context of controlling Deep-Space One, Williams and Nayak (1996, 1997) present a reactive planner, Burton, that generates a single control action for the main engine subsystem of the spacecraft, and compensates for anomalies at every step. Given a high-level goal (for example, thrust in one of the engines), Burton continually tries to transition the system toward a state that satisfies the desired goal. What is particularly relevant for us is that Burton's task can be described as a STRIPS planning problem in which each operator affects only a *single* variable (hardware component) – Williams and Nayak (1997) argue that in physical hardware it is usually the case that each state variable is commanded separately. However, Burton is based on two additional important restrictions: First, the planner is explicitly supplied with a serialization order for any satisfiable set of goal. Second, all operators must be reversible.

One of the reasons cited for designing Burton as a reactive planner that generates a single action at a time was the potential intractability of generating whole plans. Indeed,

Williams and Nayak were pessimistic about the prospects of generating whole plans quickly even for Burton, i.e., for problem instances with serializable sub-goals and single-effect operators. As our results show, this pessimism was not fully justified.

Our work continues the study of planning with unary operators. This apparently easier problem is in fact as hard as the general STRIPS planning problem (Bylander, 1994). However, we can obtain finer distinctions and some positive results if we pay closer attention to the causal structure of the domain. For example, it is easy to show that when the causal graph is a tree, it is easy to determine a serializability ordering over any set of sub-goals, and consequently, obtain a plan in polynomial time. In this paper we analyze the relationship between the domain's causal graph and the complexity of plan generation and plan existence. In particular we prove the following results:

- When the causal graph forms a polytree (the induced undirected graph is acyclic), and its node indegree is bounded by a constant, then plan existence and plan generation are polynomial.

- When the causal graph is directed-path singly connected (there is at most one directed path between any pair of nodes), then plan existence is NP-complete.

- In general, plan generation for the problems with acyclic causal graphs is provably intractable, i.e., the problem requires exponential time. The corresponding claim is derived from a previous result by Jonsson and Bäckström (1998b). However, we show that the complexity of plan generation for these problems can be bounded by a function of the number of paths within the causal graph.

Note that the complexity of the problems with polytree causal graphs but with unbounded node indegree remains an open problem – it is still to be shown whether they can be solved in polynomial time, or they are NP-complete.

Finally, we relate our results to an old open question: how difficult is it to generate plans for problems with *serializable subgoals* (Korf, 1987)? This question was stated by Bylander (1992), and different hypotheses were raised by different researchers. Here, we present a clear, though somewhat disappointing answer: First, our results suggest that even when the underlying causal graph of the problem is acyclic (and thus the problem is known to be serializable), finding a serialization ordering on the problem subgoals may be hard. Second, we show that even if the actual serialization ordering on the subgoals is known, solving the problem is not necessarily easy.

The rest of this paper is organized as follows: In Section 2 we first introduce some basic formalism used in the paper, then discuss, motivate and illustrate the notion of causal graph. In Sections 3 and 4 we present our results on the relation between the form of the causal graph and the complexity of the planning problem. In Section 5 we discuss the sub-goal serializability issue and the impact of our results on it. In Section 6 we describe some related work on complexity of planning, and connect our work with the previous results. We summarize in Section 7. Finally, Appendix A provides a short review of the POP algorithm (Penberthy & Weld, 1992), and Appendix B provides some of the proofs.

## 2. Basic Formalism and Causal Graphs

In this paper we consider only propositional planning problems, using the *propositional* STRIPS *with negative goals* formalism (Bylander, 1994), in which both positive and negative preconditions are allowed. Following Bäckström and Klein (1991b), we distinguish between preconditions and prevail conditions. In the former case the variable involved changes its value after the operator is executed, while in the latter case the value does not change. The post-condition of an operator expresses which state variables it changes and what values these variables will have after executing the operator. The pre-condition specifies which values these changed variables must have before the operator is executed. The prevail condition specifies which of the unchanged variables must have some specific value before execution of the operator and what these values are. Hence, prevail conditions, such as having a visa, are needed in order to apply an operator, such as Enter-USA, but their values do not change after the operator is applied. Finally, we assume that an operator is applicable if and only if both its pre- and prevail conditions are satisfied.

Formally, we assume that a problem instance is given by a quadruple $\Pi = \langle \mathcal{V}, \Lambda, Init, Goal \rangle$, where:

- $\mathcal{V} = \{v_1, \ldots, v_n\}$ is a set of propositional *state variables*, each one with an associated *binary domain* $\mathcal{D}(v_i)$. The domain $\mathcal{D}(v_i)$ of the variable $v_i$ induces an *extended domain* $\mathcal{D}^+(v_i) = \mathcal{D}(v_i) \cup \{u\}$, where $u$ denotes the *unspecified* value.

- $Init$ is an initial, fully specified state, i.e. $Init \in \mathcal{D}(v_1) \times \ldots \times \mathcal{D}(v_n)$.

- $Goal$ is a set of possible goal states. We assume that such a set is specified by a partial assignment on $\mathcal{V}$, thus $Goal \in \mathcal{D}^+(v_1) \times \ldots \times \mathcal{D}^+(v_n)$.

- $\Lambda = \{A_1, \ldots, A_N\}$ is a finite set of *operators* of the form $\langle \mathsf{pre}, \mathsf{post}, \mathsf{prv} \rangle$, where $\mathsf{pre}, \mathsf{post}, \mathsf{prv} \subseteq \mathcal{D}^+(v_1) \times \ldots \times \mathcal{D}^+(v_n)$ denote the pre-, post-, and prevail condition, respectively. In what follows, by $\mathsf{pre}(A)$, $\mathsf{post}(A)$, and $\mathsf{prv}(A)$ we denote the corresponding conditions of an operator $A$, and by $\mathsf{pre}(A)[i]$, $\mathsf{post}(A)[i]$, and $\mathsf{prv}(A)[i]$ the corresponding values of the variable $v_i$.

  For every $v_i \in \mathcal{V}$, we must have either $\mathsf{pre}(A)[i] = u$ or $\mathsf{prv}(A)[i] = u$. Further, $\mathsf{post}(A)[i] \neq u$ if and only if $\mathsf{pre}(A)[i] \neq u$, in which case $\mathsf{post}(A)[i] \neq \mathsf{pre}(A)[i]$.

In this paper we analyse only planning problems with *unary* operators. Therefore, in what follows, we assume that, for each operator $A \in \Lambda$, we have that:

1. there exists a variable $v_i \in \mathcal{V}$, such that $\mathsf{pre}(A)[i] \neq u$, and

2. for each other variable $v_j \in \mathcal{V} - \{v_i\}$, $\mathsf{pre}(A)[j] = u$.

Note that specifying both pre- and postconditions in case of only propositional variables is redundant, and we use it only to simplify the presentation. Likewise, our assumption that $\mathsf{post}(A) \neq u$ implies $\mathsf{pre}(A) \neq u$ is different from the usual STRIPS formalism, and requires an exponential time translation in general. However, in our case of only unary operators, this translation takes only linear time.

## 2.1 Causal Graphs

*Causal graphs* were used by Williams and Nayak (1997) as a tool for describing the structure of planning domains with unary operators. They represent a dependence relation between the state variables in the domain. A causal graph $\mathcal{G}$ is a directed graph whose nodes correspond to the state variables. An edge from $p$ to $q$ appears in the causal graph if and only if some operator that changes the value of $q$ has a prevail condition involving some value of $p$. Hence the immediate predecessors of $q$ in $\mathcal{G}$ are all those variables that affect our ability to change the value of $q$. Such a problem structure was introduced by Knoblock (1994) in the context of automatic generation of abstractions for planning. The causal graph is an intuitive model which is easily constructed given any planning problem.

Causal graphs are not the only graphical structure that can be derived from a given planning problem, and effectively exploited in solving it. For instance, graphs in which operators and literals (and not variables/propositions) are represented by the nodes, and the edges represent both prevail and preconditions were introduced by Etzioni (1993) and Smith and Peot (1993). In particular, problem space graphs of Etzioni (1993) and operator graphs of Smith and Peot (1993) were proposed as mechanisms to reduce the number of threats that arise during the total-order and partial-order planning, respectively. However, in this paper we focus on the causal graphs, since they were shown to be especially informative when all operators are unary (Jonsson & Bäckström, 1998b; Williams & Nayak, 1997).

Causal graphs have an important potential role in the design of autonomous industrial systems, as argued and demonstrated by Williams and Nayak (1997): Unary operators are natural when the manipulated objects are hardware components, since the basic control actions in such systems change the state of a single hardware component. The applicability of these control actions in any state depends on the state of the affected component as well as on the state of the related hardware components. This naturally gives rise to a planning domain with unary operators. Moreover, since the state variables correspond to hardware components, in the induced causal graph we typically see that the prevail dependencies between variables are usually implicitly entailed by the inter-composition of the hardware components. Thus, the causal graph of such domains resembles the structure of and the relationships between the system's hardware components. This resemblance has important practical ramifications for system design given the relationship between causal graph structure and the complexity of plan generation: It enables the system designer to consider the effect of his hardware design on the system's ability to autonomously generate control sequences.

A case in point is the planning problem studied by Williams and Nayak (1997), which had a number of important features: all operators were unary and reversible, and the causal graph was acyclic. Williams and Nayak argued that acyclic connectivity frequently occurs in designed systems. However, the requirement that all operators should be reversible seems to us restrictive, and it has important impact on the complexity of the problem. In the case of the Burton planner (Williams & Nayak, 1997), there were good reasons to make this assumption. Burton's reactive nature precludes extensive deliberation on the consequences of its operators. Thus it leaves open the possibility that operators may degrade the system's capabilities, leading it to dead-ends. In that case, the restriction to reversible operators was required in order to achieve a more reliable system. As we show later, in certain cases, complete plans can be generated efficiently even when the operators are not reversible.

Williams and Nayak's work has another interesting aspect, as noted by Weld (1999). For a long time, researchers have known that planning problems with serializable subgoals are likely to be easier to solve. Williams and Nayak recognized that their spacecraft configuration task was serializable (many real-world problems are not), and, more importantly, they developed a fast algorithm for computing the correct order based on the fact that the underlying causal graph is acyclic. However, their algorithm makes heavy use of the fact that all operators are reversible. Informally, reversibility implies that we can solve our subgoals one by one as long as they are consistent with some topological order of the causal graph without taking into account any global considerations: any side-effect can always be undone. Without the assumption of operator reversibility, it is relatively easy to show that Williams and Nayak's algorithm works only if the causal graph forms a directed chain. Even when the causal graph is a tree, although the problem is easy, one must take care in the choice of which subgoal to achieve next when operators are not reversible. As we show later, when the structure of the causal graph is more complicated than a directed tree then either the problem is hard or, if not, a more sophisticated algorithm is required.

Finally, we note that the existence of reversible operators might make the problem seem easier than it actually is. In this paper we present an example of a propositional planning problem with unary operators, acyclic causal graph, and totally reversible operators, the minimal solution of which is exponentially long in the size of the problem's description.

## 2.2 Example

In order to illustrate the notion of a causal graph, consider the following example, inspired by the work of Williams and Nayak (1997) on controlling the main engine subsystem of the Cassini spacecraft, in general, and its valve driver circuitry, in particular.

Each valve $VL$ (on/off) is controlled by a valve driver $VLD$ (open/close), and a safety control unit $SCU$ (safe/unsafe). Each driver controls exactly one valve, while a safety control unit can control several valves. Commands to the driver are sent via a driver control unit, that consist of two switches, $S^l$ and $S^r$, which can be either on or off. The activating states of $S^l$ and $S^r$ are described below. A valve reacts (by a state change) to a command from its driver only if (i) the instruction actually involve a state change (i.e., an open valve should not be reopened), and (ii) the safety control unit indicates that manipulating the valve is safe. In addition, the valve can be closed if the safety control unit indicates an unsafe situation. For simplicity of presentation, Table 1 presents the operator set for controlling the valves and valve drivers only. The dashed boxes stand for driver control units, two switches in each.

Now suppose that the valves $VL_1$ and $VL_2$, with the drivers $VLD_1$ and $VLD_2$, respectively, are controlled by a shared safety control unit $SCU$. Given the operator set in Table 1, the causal graph for controlling this subsystem is presented in Figure 1.

## 3. Polytree Causal Graphs

Starting at this section, we show how, by bounding the structural complexity of the causal graph, we can bound the complexity of plan generation. Recall that we use a propositional language (binary variables) to describe the state of the world, and each operator is described by its prevail conditions, single precondition, and single effect (or post-condition). The precondition and the effect are two literals, one the negation of the other.

| Affected component | pre | post | prv |
|---|---|---|---|
| $VLD$ | $close$ | $open$ | $S^l = 1 \ \wedge \ S^r = 0$ |
| | $open$ | $close$ | $S^l = 0 \ \wedge \ S^r = 1$ |
| $VL$ | $on$ | $off$ | $VLD = close \ \wedge \ SCU = safe$ |
| | $off$ | $on$ | $VLD = open \ \wedge \ SCU = safe$ |
| | $on$ | $off$ | $SCU = unsafe$ |

Table 1: A subset of the operator set for the valve circuitry controller example.

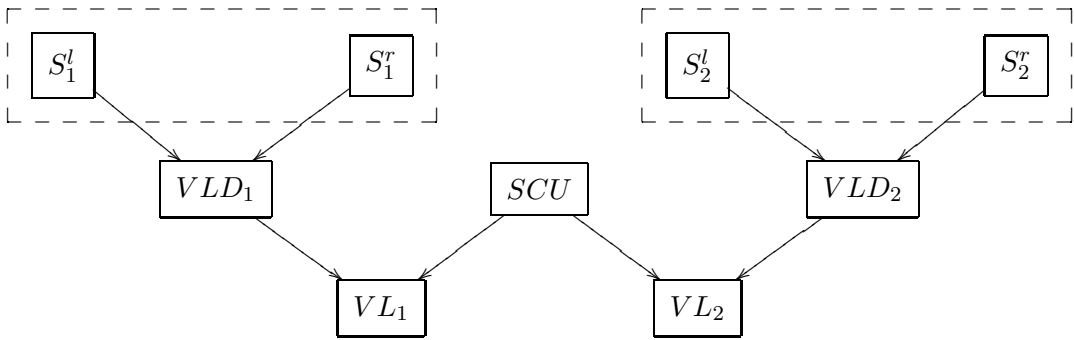

Figure 1: Causal graph for the example.

A causal graph forms a *polytree* if there is a single path between every pair of nodes in the induced *undirected* graph[1], i.e., the induced undirected graph is a tree. For example, the causal graph presented in Figure 1 forms a polytree. For this class of problems we present a planning algorithm which is polynomial if the indegree of all nodes in the causal graph is bounded by a constant. We argue that this assumption is reasonable if the prevail dependencies reflect the inter-composition of some controlled hardware components (Williams & Nayak, 1997).

Given a propositional planning instance with a polytree causal graph, we:

1. Provide a *general* upper bound for the number of times that a variable may be required to change its value on a valid, irreducible plan.

2. Using this general upper bound, provide a polynomial time procedure, called DETERMINE-MAX-SEQUENCE, that, given a variable $v$, determines the *actual* maximal number of times that $v$ can change its value on a valid, irreducible plan.

3. Provide a preprocessing algorithm that: (a) determines whether or not a plan for a given problem instance of our class exists, and (b) performs a substantial amount of preprocessing for the subsequent step of plan generating. This algorithm is based on a top-down execution of DETERMINE-MAX-SEQUENCE on the variables of the given problem instance.

---

1. These graphs are also known as *singly connected DAGs*.

4. If the answer of the plan existence check is positive we run a particular *deterministic* instance of the POP algorithm[2] (Penberthy & Weld, 1992), called POP-PCG, that generates the required plan using the information provided by the preprocessing algorithm, without backtracking, in linear time.

Informally, this process is based on the following properties of the planning problems with polytree causal graph. First, the bound achieved in step 1 is necessary for the steps 2-3, which are the main steps of our technique. By itself, this bound will be valid not only for a polytree, but for a wider class of directed-path singly connected causal graphs. However, steps 2-3 will be valid for polytree causal graphs only, because of the following properties of this form of dependence relation between the variables:

(i) Given a variable $v \in \mathcal{V}$, changing the value of a parent (immediate predecessor) $w \in \mathsf{pred}(v)$ does not require any changes of neither other parents of $v$, nor their predecessors in the causal graph.

(ii) The number of times that a variable $v$ will be able to change its value along a valid plan for a given problem instance depends directly both on these numbers for $\mathsf{pred}(v)$, and on the actual ordering of the value changes of $\mathsf{pred}(v)$.

(iii) From (i) it follows that *all* the possible orderings of the value changes of $\mathsf{pred}(v)$ are legal. In addition, it will be shown that chosing an ordering for the value changes of $\mathsf{pred}(v)$ will not affect our ability to change the value of any variable except of $v$.

(iv) The crucial part of the process (steps 2-3) is basically about finding the *right* ordering of the *right* number of value changes of $\mathsf{pred}(v)$ for each variable $v \in \mathcal{V}$. By synchronizing these changes to $v$'s parents appropriately, we can increase the number of possible changes to $v$.

We start with some notation. First, a valid plan $P$ for a given planning instance $\Pi$ will be called *irreducible* if any subplan $P'$ of $P$ is not a plan for $\Pi$, in the following sense: Removal of any subset of (not necessarily subsequent) actions from $P$ makes the resulting plan either illegal, or its initial state is not $Init$, or its goal state is not one of the states specified by $Goal$. The notion of irreducible plans was introduced by Kambhampati (1995), where it was exploited for admissible pruning of partial plans during search[3].

---

2. A short review of the POP algorithm, and the corresponding formalism is provided in Appendix A. For those familiar with the algorithm, we note one slight technical change, stemming from the use of unary operators. POP uses two fictitious actions $A_0$ and $A_\infty$ to capture the initial and goal state, respectively. Here, we replace each of these actions by a set of actions, each with a single effect. The (fictitious) action setting the initial value of variable $v_i$ is denoted $A_i^0$ and the fictitious action whose precondition is the goal value of variable $v_i$ is denoted $A_i^*$.

3. Irreducible plans were called in (Kambhampati, 1995) *minimal plans.* However, we decided to change the name of this concept in order to prevent an ambiguity between "minimal as irreducible" and "minimal as optimal".

Now, given a planning instance $\Pi$, let $\mathcal{P}$ be the set of *all* irreducible plans for $\Pi$. We denote by $\mathsf{MaxReq}(v)$ the maximal number of times that a variable $v \in \mathcal{V}$ changes its value in the course of execution of an irreducible plan for $\Pi$. Formally, let $\mathsf{Req}(P, v)$ be the number of times that $v$ changes its value in the course of execution of a plan $P$. Then,

$$\mathsf{MaxReq}(v) = \max_{P \in \mathcal{P}} \{\mathsf{Req}(P, v)\}$$

Observe that, for any planning problem with unary operators, a variable must change its value at most once for each required change of its immediate successors in the causal graph (in order to satisfy the necessary prevail conditions), and then at most once in order to obtain the value requested by the goal. Thus, for all variables in $\mathcal{V}$, $\mathsf{MaxReq}(v)$ satisfies:

$$\mathsf{MaxReq}(v) \leq 1 + \sum_{succ(v)} \mathsf{MaxReq}(u) \tag{1}$$

where $\mathsf{succ}(v)$ denotes the immediate successors of $v$ in the corresponding causal graph. Adopting the terminology from (Domshlak & Shimony, 2003; Shimony & Domshlak, 2002), a directed acyclic graph $G$ is *directed-path singly connected* if, for every pair of nodes $s, t \in G$, there is at most one directed path from $s$ to $t$. The following lemma shows that if the causal graph forms a directed-path singly connected DAG then we can bound $\mathsf{MaxReq}(v)$ by $n$. Clearly, all polytrees are directed-path singly connected DAGs, but not vice versa.

**Lemma 1** *For any solvable problem instance $\Pi$ with a directed-path singly connected causal graph over $n$ variables, for any variable $v$, we have that $\mathsf{MaxReq}(v) \leq n$.*

**Proof:** The proof is by induction on $n$. For $n = 1$ it is obvious that $\mathsf{MaxReq}(v) \leq 1$. Now suppose that when $|\mathcal{V}| = n - 1$ then for any $v \in \mathcal{V}$,

$$\mathsf{MaxReq}(v) \leq n - 1$$

Let $\Pi'$ be some problem instance for which $|\mathcal{V}'| = n$. Suppose that the variables in $\mathcal{V}' = \{v_1, \ldots v_n\}$ are topologically ordered based on the domain's causal graph. Clearly, $v_n$ is a leaf node (i.e., $\mathsf{succ}\ (v_n) = \emptyset$). We will denote by $\Pi$ the problem instance obtained by removing $v_n$ from the domain, and the corresponding variable set by $\mathcal{V}$. According to Eq. 1, for each immediate predecessor $v$ of $v_n$ in the causal graph,

$$\mathsf{newMaxReq}(v) \leq \mathsf{MaxReq}(v) + \mathsf{newMaxReq}(v_n) \leq \mathsf{MaxReq}(v) + 1$$

where $\mathsf{newMaxReq}(v)$ denotes $\mathsf{MaxReq}(v)$ with respect to $\Pi'$. Generally, since the causal graph is directed-path singly connected, for each variable $v \in \mathcal{V}'$,

$$\mathsf{newMaxReq}(v) \leq \begin{cases} \mathsf{MaxReq}(v) + 1, & \text{if there is a path from } v \text{ to } v_n \\ \mathsf{MaxReq}(v), & \text{otherwise} \end{cases} \tag{2}$$

and thus, for each $v \in \mathcal{V}'$, holds

$$\mathsf{newMaxReq}(v) \leq n$$

$\square$

Recall that $\mathsf{MaxReq}(v)$ stands for an upper bound on the number of value changes of $v$ that may be required by a valid, irreducible plan. However, the maximal achievable number of value changes of $v$, denoted by $\mathsf{MaxPoss}(v)$ can be greater or less than $\mathsf{MaxReq}(v)$. For example, if $v$ has no predecessors in the causal graph, and there are two operators affecting $v$ differently, then $\mathsf{MaxPoss}(v) = \infty$.

We denote the upper bound on the *feasible* number of value changes of $v$ that may be required in a valid, irreducible plan for $\Pi$ by $\mathsf{FMaxReq}(v)$. Informally, no more than $\mathsf{MaxPoss}(v)$ value changes of $v$ *can* be required and no more than $\mathsf{MaxReq}(v)$ value changes of $v$ *should* be required, thus

$$\mathsf{FMaxReq}(v) = \min(\mathsf{MaxPoss}(v), \mathsf{MaxReq}(v)) \tag{3}$$

Determining $\mathsf{FMaxReq}(v)$ for all variables requires explicit examination of a given problem instance. Recall that here we restrict the causal graph of $\Pi$ to form a polytree. To simplify the presentation, we assume that the goal values are specified for all state variables, i.e. $Goal \in \mathcal{D}(v_1) \times \ldots \times \mathcal{D}(v_n)$. Later we show that this assumption does not affect the generality of the algorithm. Denote by $v^0$ and $v^*$ the initial and the goal values of $v$ in $\Pi$, and by $\Lambda_v \subseteq \Lambda$ the set of all operators affecting $v$. First we examine the root variables of the causal graph, then we analyze the rest of the variables.

Denote by $\mathsf{pred}(v)$ the immediate predecessors of $v$ in the causal graph. If $\mathsf{pred}(v) = \emptyset$, then there are at most two operators $A_v^-, A_v^+$ in $\Lambda_v$: $A_v^+$ has $v^*$ as its postcondition, while $A_v^-$ has the reverse effect. Since these operators have no prevail condition, if both $A_v^-$ and $A_v^+$ are presented in $\Lambda$, then they can be applied one after another an infinite number of times. Therefore, from Eq. 3, $\mathsf{FMaxReq}(v) = n$. If $\Lambda_v \neq \{A_v^-, A_v^+\}$ then we have two cases: If the initial and the goal values of $v$ are the same, then we cannot change the value of $v$ and reconstruct it later, and thus $\mathsf{FMaxReq}(v) = 0$. Alternatively, if the initial and the goal values of $v$ are different then if $\Lambda_v = \{A_v^+\}$ then we can achieve the goal value of $v$ but only once and thus $\mathsf{FMaxReq}(v) = 1$. Otherwise, the goal value of $v$ is unachievable, thus the given problem instance is unsolvable. Table 2 summarize this analysis.

|  | $\Lambda_v$ | $\mathsf{FMaxReq}(v)$ |
|---|---|---|
| $v^0 = v^*$ | $\{A_v^-, A_v^+\}$ | n |
|  | otherwise | 0 |
| $v^0 \neq v^*$ | $\{A_v^-, A_v^+\}$ | n |
|  | $\{A_v^+\}$ | 1 |
|  | otherwise | no solution |

Table 2: $\mathsf{FMaxReq}(v)$ values for the root variables in the causal graph.

Now consider a variable $v$ which is presented by an internal node in the causal graph: $\mathsf{pred}(v) = \{w_1, \ldots, w_k\} \neq \emptyset$. Observe that the number of possible value changes of $v$ depends on and only on:

1. The initial and the goal values of $v$, i.e. $v^0$ and $v^*$.

2. The set of operators affecting $v$, i.e., $\Lambda_v$.

3. The maximally possible (but still reasonable) number of times that predecessors of $v$ can change their values, i.e., $\mathsf{FMaxReq}(w_1), \ldots, \mathsf{FMaxReq}(w_k)$.

4. The actual scheduling of the value changes of the predecessors of $v$.

The last point is crucial – it means that in order to determine $\mathsf{FMaxReq}(v)$ we should find a particular scheduling of the value changes of $\mathsf{pred}(v)$ that allows such a maximal number of value changes for $v$. The corresponding interleaving sequence of $v$'s values, starting and finishing by $v^0$ and $v^*$ respectively, with $\mathsf{FMaxReq}(v)$ value changes will be called *maximal* and will be denoted by $\sigma(v)$ ($|\sigma(v)| = \mathsf{FMaxReq}(v) + 1$).

From Lemma 1, for $1 \leq i \leq k$, we have $\mathsf{FMaxReq}(w_i) \leq n$, thus the number of different orderings of value changes of $\mathsf{pred}(v)$ can be exponential in $n$. For instance, when, for $1 \leq i \leq k$, we have $\mathsf{FMaxReq}(w_i) = n$, this number of different orderings can be expressed as:

$$\prod_{i=1}^{k-1} \sum_{j=1}^{n} \binom{n-1}{j-1} \binom{ni+1}{j} \gg 2^{nk}$$

where the correctness of the expression on the left side of the inequality is shown by Lemma 4 (see Appendix B, p. 347). Clearly, we cannot check all these orderings in a naive manner. Following, we provide an algorithm that determines $\sigma(v)$ in time which is polynomial in $n$.

For clarity of presentation we want to distinguish between the different elements of a maximal sequence $\sigma(v)$. Since all variables are binary, we denote the initial value of $v$, $v^0$, by $\mathsf{b}_v$ and the opposite value by $\mathsf{w}_v$ (black/white). Similarly, $\mathsf{b}_i$ and $\mathsf{w}_i$ will stand for the corresponding values of the variable $v_i$. If so, we can think about all the operators in $\Lambda$ as described in this language. Likewise, we sequentially number the appearances of each value of $v$ on $\sigma(v)$. For example, $\mathsf{b}_v^i$ stands for the $i$th appearance of the value $\mathsf{b}_v$ along $\sigma(v)$. To illustrate this notation, suppose that $\mathcal{D}(v) = \{true, false\}$, the initial value of $v$ is $v^0 = true$, and $\mathsf{FMaxReq}(v) = 4$. Then, we have:

$$\begin{aligned}
\mathsf{b}_v &\equiv true \\
\mathsf{w}_v &\equiv false \\
\sigma(v) &= \mathsf{b}_v^1 \cdot \mathsf{w}_v^1 \cdot \mathsf{b}_v^2 \cdot \mathsf{w}_v^2 \cdot \mathsf{b}_v^3
\end{aligned}$$

First, for every variable $v$, every operator $A \in \Lambda_v$ is extended to a set of operators that explicitly specify prevail values for *all* parents of $v$ in the causal graph: If $|\mathsf{pred}(v)| = k$, and the prevail condition of $A$ is specified only in terms of some $0 \leq k' \leq k$ parents[4] of $v$, then $A$ is extended to a set of $2^{k-k'}$ operators, where each operator extends $A$ by an instantiation of the previously unspecified parents of $v$. For example, consider a variable $v$ with $\mathsf{pred}(v) = \{u, w\}$, and an operator

$$A = \{\mathsf{pre} : \{\mathsf{b}_v\}, \ \mathsf{post} : \{\mathsf{w}_v\}, \ \mathsf{prv} : \{\mathsf{b}_u\}\},$$

---

4. For every other parent $w_j$ of $v$, we have $\mathsf{prv}(A)[j] = \mathsf{u}$.

the prevail condition of which does not involve $w$. This operator is extended to a pair of operators:

$$A' = \{\mathsf{pre} : \{\mathsf{b}_v\}, \ \mathsf{post} : \{\mathsf{w}_v\}, \ \mathsf{prv} : \{\mathsf{b}_u, \mathsf{b}_w\}\}$$
$$A'' = \{\mathsf{pre} : \{\mathsf{b}_v\}, \ \mathsf{post} : \{\mathsf{w}_v\}, \ \mathsf{prv} : \{\mathsf{b}_u, \mathsf{w}_w\}\}$$

corresponding to the possible values of $w$. In what follows, we refer to the operator set resulting from such a compilation of $\Lambda$ as $\Lambda^{\bowtie}$. Note that, under the assumption of constantly bounded maximal indegree $\kappa$ of the causal graph, compiling $\Lambda$ into $\Lambda^{\bowtie}$ takes only polynomial time, since, for every variable $v$, $|\Lambda_v^{\bowtie}| \leq 2^{\kappa+1}$, and thus $|\Lambda^{\bowtie}| = O(n2^{\kappa+1})$.

Given the maximal sequences $\sigma(w_1), \dots, \sigma(w_k)$ and the operator set $\Lambda_v^{\bowtie}$ we construct a directed graph (denoted as $G'_e(v)$) that captures *all (and only) feasible* sequences of, up to $n$, value changes of $v$, where each value change is annotated with the corresponding assignment on $\mathsf{pred}(v)$. Although the number of the captured sequences can be exponential in $n$, the size of $G'_e(v)$ is polynomial in $n$. With respect to this graph, the problem of finding the maximal sequence $\sigma(v)$ is reduced to the problem of finding a longest path from a given node to an arbitrary other node in a directed acyclic graph.

The graph $G'_e(v)$ is created in three incremental steps. At the first step, given the maximal sequences $\sigma(w_1), \dots, \sigma(w_k)$ and the operator set $\Lambda_v^{\bowtie}$ we construct a directed labeled graph $G(v)$ capturing information about all sequences of assignments on $\mathsf{pred}(v)$ that can enable $n$ or less value flips of $v$. The graph $G(v)$ is defined as follows:

1. $G(v)$ consist of $\eta$ nodes, where

$$\eta = \begin{cases} n, & ((n = 2j) \text{ and } (v^0 = v^*)) \text{ or} \\ & ((n = 2j + 1) \text{ and } (v^0 \neq v^*)), \ j \in \mathbb{N} \\ n-1, & \text{otherwise} \end{cases}$$

2. $G(v)$ forms a *2-colored multichain*, i.e., (i) the nodes of the graph are colored by black and white, starting by black; (ii) there are no two subsequent nodes with the same color; (iii) for $1 \leq i \leq \eta - 1$, edges from the node $i$ are only to the node $i + 1$.

   Observe that such a construction of $G(v)$ promises that the color of the last node will be consistent with $v^*$.

3. The nodes of $G(v)$ are denoted precisely by the elements of the maximal sequence $\sigma(v)$, i.e., $\mathsf{b}_v^i$ stands for the $i$th black node in $G(v)$.

4. Suppose that there are $m$ operators in $\Lambda_v^{\bowtie}$ that change the value of $v$ from $\mathsf{b}_v$ to $\mathsf{w}_v$. In this case, for each $i$, there are $m$ edges from $\mathsf{b}_v^i$ to $\mathsf{w}_v^i$, and $|\Lambda_v^{\bowtie}| - m$ edges from $\mathsf{w}_v^i$ to $\mathsf{b}_v^{i+1}$. All edges are labeled by the prevail conditions of the corresponding operators, i.e., a $k$-tuple of the values of $w_1, \dots, w_k$. This tuple is denoted by $l(e)$ (label of the edge $e$) and its component, corresponding to a predecessor $w_i$, is denoted by $l(e)_{w_i}$.

This formal definition of $G(v)$ is relatively complicated, thus we provide a demonstrating example: Suppose that we are given a problem instance over 5 variables, and we consider a variable $v$ with $\mathsf{pred}(v) = \{u, w\}$, $v^0 = \mathsf{b}_v$, and $v^* = \mathsf{w}_v$. Recall that every operator in $\Lambda^{\bowtie}$ is presented as a three-tuple $\langle \mathsf{pre}, \mathsf{post}, \mathsf{prv} \rangle$ of pre-, post-, and prevail conditions of the operator respectively. Suppose that:

$$\sigma(u) = \mathsf{b}_u^1 \cdot \mathsf{w}_u^1 \qquad \sigma(w) = \mathsf{b}_w^1 \cdot \mathsf{w}_w^1 \cdot \mathsf{b}_w^2 \cdot \mathsf{w}_w^2$$

$$\Lambda_v^{\bowtie} = \left\{ \begin{array}{l} A_v^1 = \{\mathsf{pre} : \{\mathsf{b}_v\}, \ \mathsf{post} : \{\mathsf{w}_v\}, \ \mathsf{prv} : \{\mathsf{b}_u, \mathsf{w}_w\}\} \\ A_v^2 = \{\mathsf{pre} : \{\mathsf{w}_v\}, \ \mathsf{post} : \{\mathsf{b}_v\}, \ \mathsf{prv} : \{\mathsf{b}_u, \mathsf{b}_w\}\} \\ A_v^3 = \{\mathsf{pre} : \{\mathsf{w}_v\}, \ \mathsf{post} : \{\mathsf{b}_v\}, \ \mathsf{prv} : \{\mathsf{w}_u, \mathsf{w}_w\}\} \end{array} \right.$$

For this case, the graph $G(v)$ is presented by Figure 2.

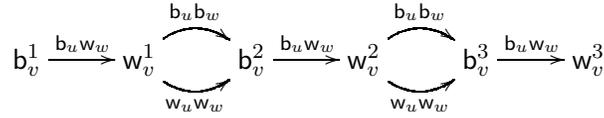

Figure 2: Example of the graph $G(v)$.

The constructed graph $G(v)$ captures information about all *potentially possible* executions of the operators in $\Lambda_v^{\bowtie}$ that can provide us $\mathsf{MaxReq}(v)$ or less value changes of $v$. Each path, started at the source node of $G(v)$, uniquely corresponds to such an execution. Although the number of these alternative executions may be exponential in $n$, this graphical representation is compact: the number of edges in $G(v)$ is $O(n \cdot |\Lambda_v^{\bowtie}|)$. Note that the information about the number of times that each operator in $\Lambda_v^{\bowtie}$ can be executed is not captured by $G(v)$. The following two steps add this information indirectly and exploit it to find a maximal sequence $\sigma(v)$.

At the second step of construction, we expand $G(v)$ with respect to the maximal sequences $\sigma(w_1), \ldots, \sigma(w_k)$ as follows: Each edge $e \in G(v)$ (which by definition corresponds to some operator $A \in \Lambda_v^{\bowtie}$), is replaced by a set of edges such that their labels correspond to all possible assignments of the elements of $\sigma(w_1), \ldots, \sigma(w_k)$ to $l(e)$ (i.e., $\mathsf{prv}(A)$). Likewise, we add a dummy source node $s_v$, with an edge from $s_v$ to the original source node of $G(v)$ labeled by a tuple of the first elements of $\sigma(w_1), \ldots, \sigma(w_k)$ (= initial values of $w_1, \ldots, w_k$). Similarly, we add a dummy target node $t_v$, with an edge from the original target node of $G(v)$ to $t_v$ labeled by a tuple of the last elements of $\sigma(w_1), \ldots, \sigma(w_k)$ (= goal values of $w_1, \ldots, w_k$). We denote this extended graph by $G'(v)$, and Figure 3 illustrates $G'(v)$ for the example above.

The extended graph $G'(v)$ can be viewed as a projection of the maximal sequences $\sigma(w_i)$, $1 \leq i \leq k$, on the graph $G(v)$. Each edge in $G(v)$ may be replaced by $O(n^k)$ edges in $G'(v)$, and thus the number of edges in $G'(v)$ is $O(n^{k+1} \cdot |\Lambda_v^{\bowtie}|)$.

It is easy to see that not all paths in $G'(v)$ starting at $s_v$ are relevant. For example, in $G'(v)$ above, an operator instance prevailed by $\mathsf{b}_u^1 \mathsf{b}_w^2$ can not be performed after an operator instance prevailed by $\mathsf{b}_u^1 \mathsf{w}_w^2$. Thus, now we are faced with the problem of finding a longest *feasible path* from $s_v$ to a node in $G'(v)$, the label of which is consistent with $v^*$. The following (last) step provides a reduction of the problem of finding a longest feasible path from $s_v$ to a $v^*$-colored node in $G'(v)$ to a known problem of finding a longest path in a directed acyclic graph. Let the graph $G'_e(v)$ have the *edges* of $G'(v)$ as nodes, and let its edges be defined by all *allowed* pairs of immediately subsequent edges in $G'(v)$: $(e, e')$ is allowed if, for $1 \leq i \leq k$, either $l(e)_{w_i} = l(e')_{w_i}$ or $l(e)_{w_i}$ appears after $l(e')_{w_i}$ on $\sigma(w_i)$. Such a construction is a variant of a so called "edge graph" known in graph theory; the addition in our case is the exclusion of non-allowed edges from it. Clearly, $G'_e(v)$ can be constructed in time polynomial in size of $G'(v)$, and the number of edges in $G'_e(v)$ is $O(n^{2k+2} \cdot |\Lambda_v^{\bowtie}|^2)$.

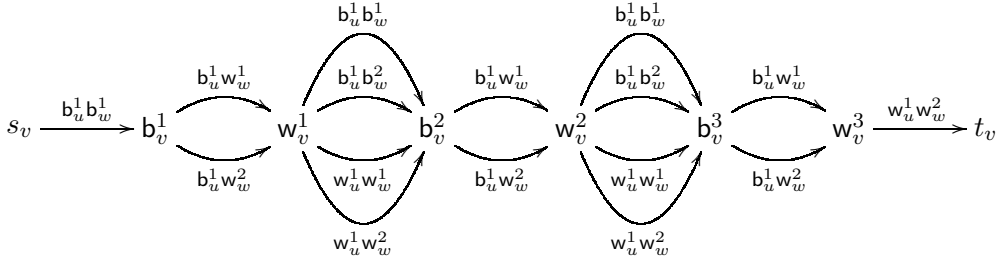

Figure 3: Example of the graph $G'(v)$.

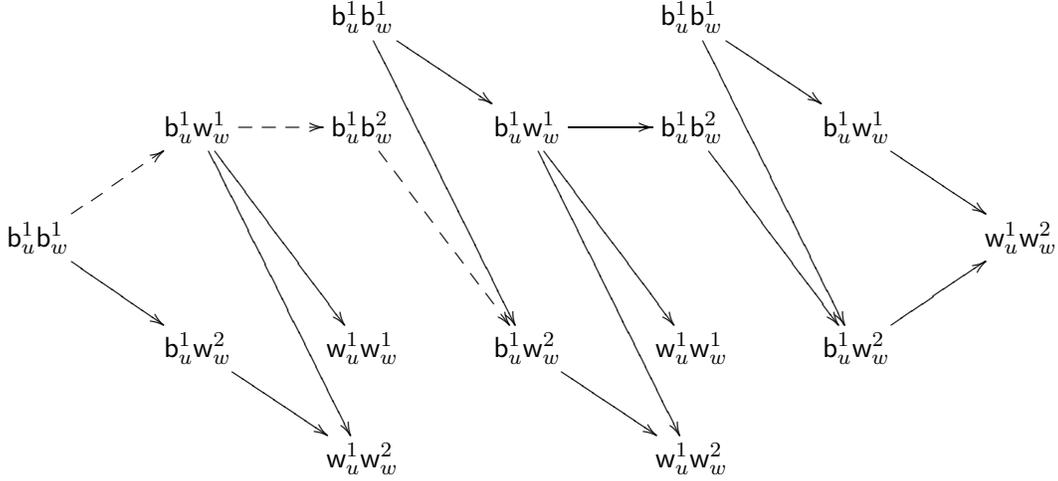

Figure 4: Example of the graph $G'_e(v)$.

Figure 4 presents $G'_e(v)$ for our example. The dashed edges present the longest path from the dummy source node to a node that corresponds to a value change from $\neg v^*$ to $v^*$ (from $\mathsf{b}_v$ to $\mathsf{w}_v$). Such a longest path in $G'_e(v)$ describes a maximal sequence of value changes $\sigma(v)$, and its length is actually $\mathsf{FMaxReq}(v) + 1$. In our example, $\sigma(v) = \mathsf{b}_v^1 \cdot \mathsf{w}_v^1 \cdot \mathsf{b}_v^2 \cdot \mathsf{w}_v^2$, and $\mathsf{FMaxReq}(v) = 3$. Note that if $v^0 = v^*$ then the empty path will be also acceptable since, in general, $v$ does not have to change its value. In this case $\mathsf{FMaxReq}(v) = 0$ and $\sigma(v)$ will consist of only one element which corresponds to the initial (= goal) value of $v$.

Observe that a longest path in $G'_e(v)$ describes not only $\sigma(v)$ but also the actual sequence of invocations of the operators from $\Lambda_v^{\bowtie}$ that provides $\sigma(v)$. We denote by $\{A(\mathsf{b}_v^j)\}$ and $\{A(\mathsf{w}_v^j)\}$ the sequences of operator instances that have as effects the corresponding elements from the sequences $\{\mathsf{b}_v^j\}$ and $\{\mathsf{w}_v^j\}$ ($\{\mathsf{b}_v^j\} \cup \{\mathsf{w}_v^j\} = \sigma(v)$) of $v$'s values, respectively. In what follows, we address these sequences of operator instances as one sequence of operator

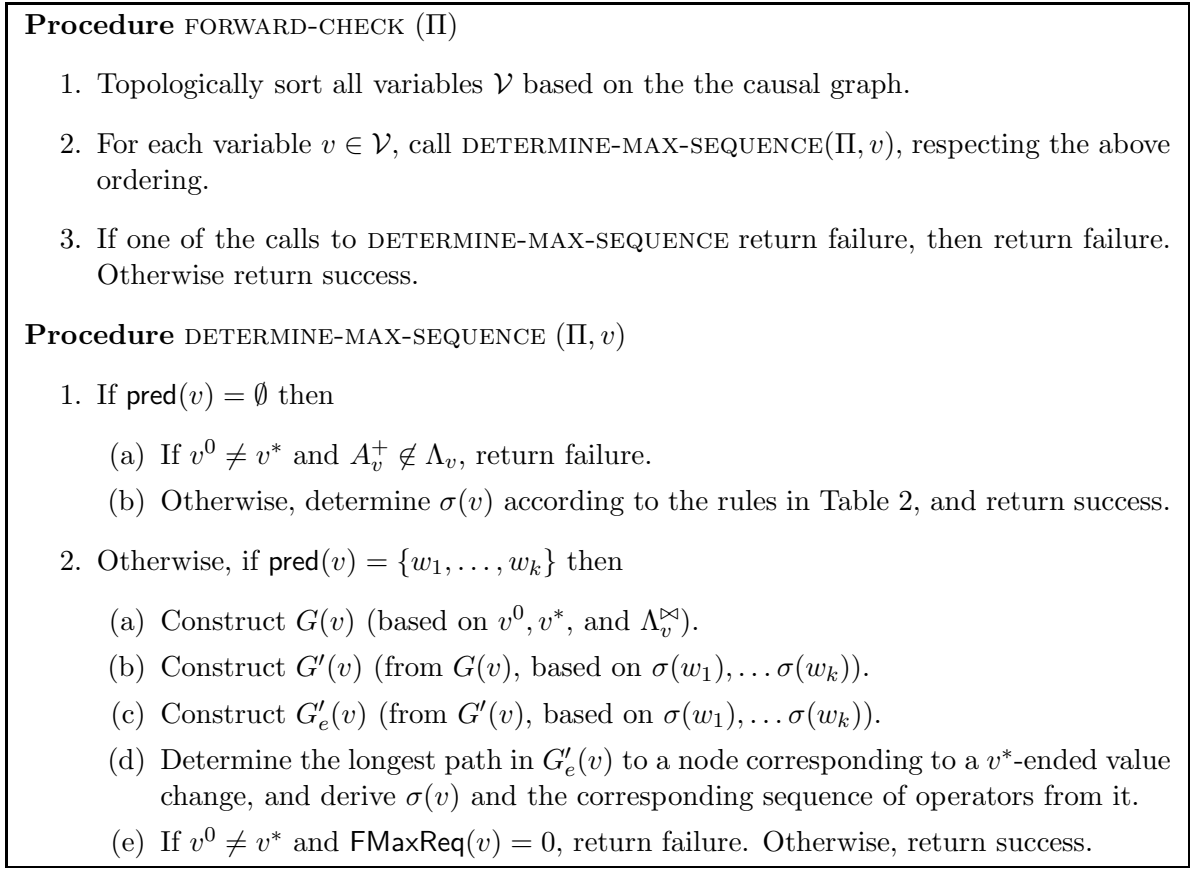

**Procedure** FORWARD-CHECK ($\Pi$)

1. Topologically sort all variables $\mathcal{V}$ based on the the causal graph.

2. For each variable $v \in \mathcal{V}$, call DETERMINE-MAX-SEQUENCE($\Pi, v$), respecting the above ordering.

3. If one of the calls to DETERMINE-MAX-SEQUENCE return failure, then return failure. Otherwise return success.

**Procedure** DETERMINE-MAX-SEQUENCE ($\Pi, v$)

1. If $\mathsf{pred}(v) = \emptyset$ then

   (a) If $v^0 \neq v^*$ and $A_v^+ \notin \Lambda_v$, return failure.

   (b) Otherwise, determine $\sigma(v)$ according to the rules in Table 2, and return success.

2. Otherwise, if $\mathsf{pred}(v) = \{w_1, \ldots, w_k\}$ then

   (a) Construct $G(v)$ (based on $v^0, v^*$, and $\Lambda_v^{\bowtie}$).

   (b) Construct $G'(v)$ (from $G(v)$, based on $\sigma(w_1), \ldots \sigma(w_k)$).

   (c) Construct $G'_e(v)$ (from $G'(v)$, based on $\sigma(w_1), \ldots \sigma(w_k)$).

   (d) Determine the longest path in $G'_e(v)$ to a node corresponding to a $v^*$-ended value change, and derive $\sigma(v)$ and the corresponding sequence of operators from it.

   (e) If $v^0 \neq v^*$ and $\mathsf{FMaxReq}(v) = 0$, return failure. Otherwise, return success.

Figure 5: FORWARD-CHECK algorithm

instances $\Gamma_v = \{A(\nu_v^i)\}_{i=2}^{\mathsf{FMaxReq}(v)}$, where $A(\nu_v^i)$ has $\nu_v^i$ as its effect, and

$$\nu_v^i = \begin{cases} \mathsf{b}_v^{\frac{i+1}{2}}, & i = 2k+1 \\ \mathsf{w}_v^{\frac{i}{2}}, & i = 2k \end{cases} \quad k \in \mathbb{N}$$

Procedure FORWARD-CHECK in Figure 5 summarizes the presented approach. Note that finding a set of longest paths from a node to all other nodes in a directed acyclic graph can be done in time linear in the size of the graph (Wiest & Levy, 1969). Therefore, the time complexity of a call to the DETERMINE-MAX-SEQUENCE procedure with a variable $v$ is bounded by the size of the constructed graph $G'_e(v)$ and thus is $O(n^{2k+2} \cdot |\Lambda_v^{\bowtie}|^2)$. FORWARD-CHECK calls DETERMINE-MAX-SEQUENCE $n$ times. Therefore, if the maximal node indegree is bounded by a constant $\kappa$, then the overall complexity of the algorithm is $O(|\mathcal{V}|^{2\kappa+3} \cdot 2^{2\kappa+2})$, i.e., polynomial in the size of the problem description.

**Theorem 1** *A given problem instance with a polytree causal graph is solvable if and only if, for each $v \in \mathcal{V}$,* FORWARD-CHECK *succeeds in constructing the maximal sequence $\sigma(v)$.*

FORWARD-CHECK fails if and only if at least one of the calls to the DETERMINE-MAX-SEQUENCE procedure fails. In turn, a call to DETERMINE-MAX-SEQUENCE on a variable $v$

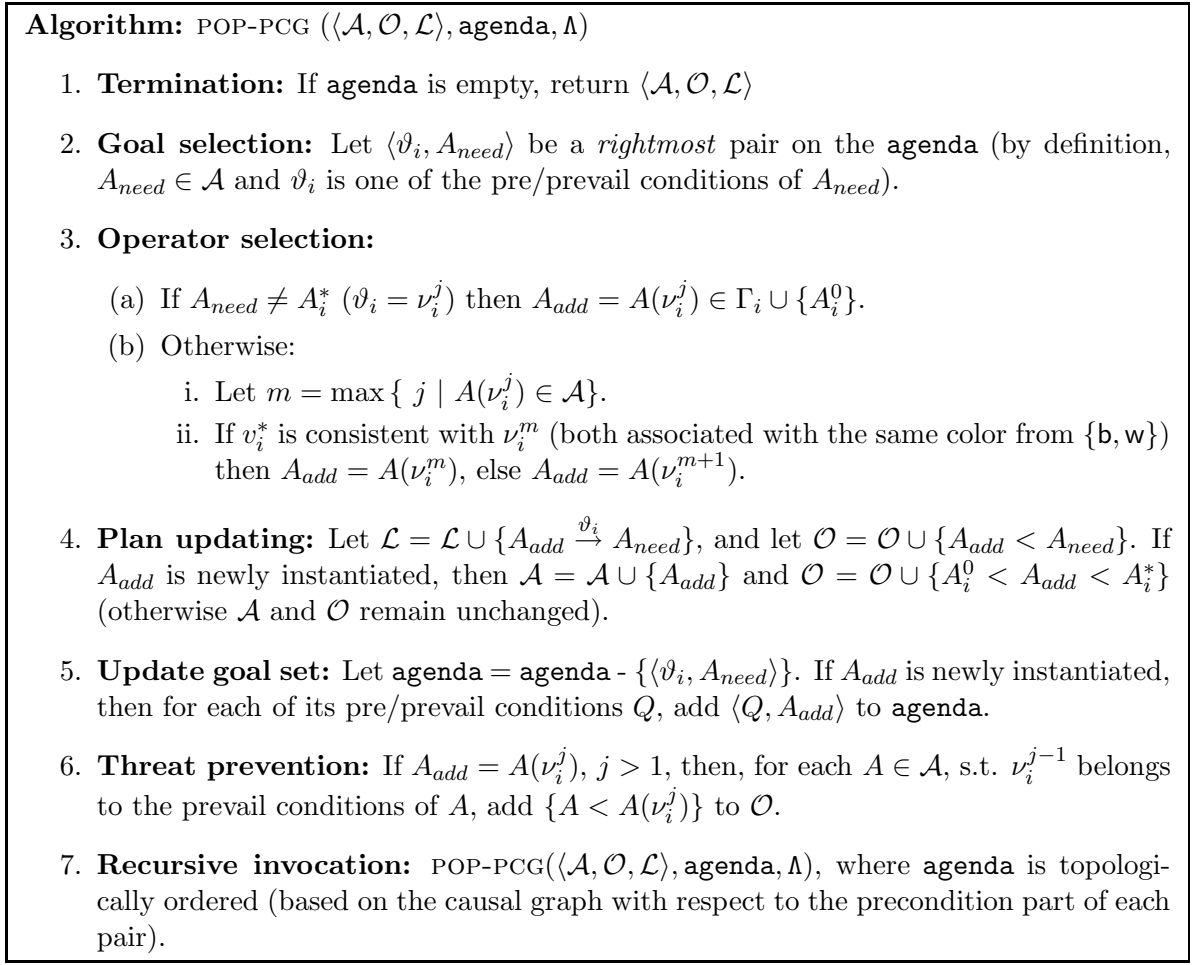

**Algorithm:** POP-PCG ($\langle \mathcal{A}, \mathcal{O}, \mathcal{L} \rangle$, agenda, $\Lambda$)

1. **Termination:** If agenda is empty, return $\langle \mathcal{A}, \mathcal{O}, \mathcal{L} \rangle$

2. **Goal selection:** Let $\langle \vartheta_i, A_{need} \rangle$ be a *rightmost* pair on the agenda (by definition, $A_{need} \in \mathcal{A}$ and $\vartheta_i$ is one of the pre/prevail conditions of $A_{need}$).

3. **Operator selection:**

   (a) If $A_{need} \neq A_i^*$ ($\vartheta_i = \nu_i^j$) then $A_{add} = A(\nu_i^j) \in \Gamma_i \cup \{A_i^0\}$.

   (b) Otherwise:

      i. Let $m = \max \{\, j \mid A(\nu_i^j) \in \mathcal{A} \}$.
      ii. If $v_i^*$ is consistent with $\nu_i^m$ (both associated with the same color from $\{\mathtt{b}, \mathtt{w}\}$) then $A_{add} = A(\nu_i^m)$, else $A_{add} = A(\nu_i^{m+1})$.

4. **Plan updating:** Let $\mathcal{L} = \mathcal{L} \cup \{A_{add} \xrightarrow{\vartheta_i} A_{need}\}$, and let $\mathcal{O} = \mathcal{O} \cup \{A_{add} < A_{need}\}$. If $A_{add}$ is newly instantiated, then $\mathcal{A} = \mathcal{A} \cup \{A_{add}\}$ and $\mathcal{O} = \mathcal{O} \cup \{A_i^0 < A_{add} < A_i^*\}$ (otherwise $\mathcal{A}$ and $\mathcal{O}$ remain unchanged).

5. **Update goal set:** Let agenda = agenda - $\{\langle \vartheta_i, A_{need} \rangle\}$. If $A_{add}$ is newly instantiated, then for each of its pre/prevail conditions $Q$, add $\langle Q, A_{add} \rangle$ to agenda.

6. **Threat prevention:** If $A_{add} = A(\nu_i^j)$, $j > 1$, then, for each $A \in \mathcal{A}$, s.t. $\nu_i^{j-1}$ belongs to the prevail conditions of $A$, add $\{A < A(\nu_i^j)\}$ to $\mathcal{O}$.

7. **Recursive invocation:** POP-PCG($\langle \mathcal{A}, \mathcal{O}, \mathcal{L} \rangle$, agenda, $\Lambda$), where agenda is topologically ordered (based on the causal graph with respect to the precondition part of each pair).

Figure 6: POP-PCG algorithm

fails if and only if the initial and the goal values of $v$ are different but there is no way to change the value of $v$ even once. Thus, if FORWARD-CHECK fails, then no plan exists.

To prove the opposite direction we proceed as follows: We define the POP-PCG algorithm (POP for polytree causal graphs) and show that it will succeed without backtracking if FORWARD-CHECK succeeds.[5] POP-PCG is described in detail in Figure 6, and it works as follows: First, let us expand each sequence of operator instances $\Gamma_i$ by $A(\nu_i^1)$ ($A(\mathtt{b}_i^1)$) which will stand for the dummy operator $A_i^0$. (Recall that up until now, only operators of the form $A(\nu_i^j)$ for $j > 1$ were defined.) The algorithm maintains a goal agenda sorted based on the causal graph structure: parent variables appear after their descendents. At each point, the next agenda item is selected; if it requires achieving some value for $v_i$ we add the corresponding operator to the plan with the desired effect (step 3a). Actually, if we would be ready to accept plans with possible redundant steps, we can omit the next step 3b from the algorithm by assuming that the goal value of each variable $v$ is the last element of the maximal sequence $\sigma(v)$. However, if we would like our plan to be irreducible, then a careful decision about the really required number of value changes of each variable is required. This decision is captured in step 3b by analysis of the value changes of a variable $v_i$ that were found necessary in the previous iterations of the algorithm in order to satisfy the predecessors of $v_i$ in the causal graph. Note that the agenda is sorted with respect to some reverse topological ordering of the causal graph, thus if an operator affecting $v_i$ was selected from the agenda then no operator affecting some predecessor of $v_i$ in the causal graph will appear on the agenda until the end of the algorithm. No threats arise in POP-PCG, and the ordering constraints are consistent.

---

[5]. For a short review of the POP algorithm, the corresponding formalism, and the description of the initial call to the algorithm, we refer the reader to Appendix A.

**Lemma 2** *If* FORWARD-CHECK *was successful then* POP-PCG *will return a valid plan.*

**Proof:**  The lemma will follow from the following claims:

1. For every agenda item, there exists an operator that has it as an effect.

2. There are no threats in the output of POP-PCG.

3. The ordering constraints in $\mathcal{O}$ are consistent.

4. The agenda will be empty after a polynomial number of steps.

For the proof see Appendix B, p. 343.  □

Recall that, for simplicity of presentation, we assumed that the goal values are specified for all state variables (single goal state), i.e. $Goal \in \mathcal{D}(v_1) \times \ldots \times \mathcal{D}(v_n)$. Now we show that the presented approach, with minor modifications, works for a set of possible goal states as well, if such a set is specified by a partial assignment on $\mathcal{V}$, i.e. $Goal \in \mathcal{D}^+(v_1) \times \ldots \times \mathcal{D}^+(v_n)$. Note that the latter assumption is widely accepted in the planning literature.

First, no modifications should be done in processing variables that are specified by $Goal$. Now, for each variable $v$, such that $v^*$ is not specified by $Goal$, the modifications are as follows:

1. The graph $G(v)$ will consist of exactly $n$ nodes. This is correct since (i) according to Lemma 1, $n$ changes of $v$ have to be sufficient, and (ii) any value change of $v$ can be its last value change.

2. No changes in construction of $G'(v)$ and $G'_e(v)$.

3. In the procedure DETERMINE-MAX-SEQUENCE:

   (a) In step 2d, determine the longest path from the dummy source node to *any* other node in the graph.

   (b) In step 2e, *always* return success.

   Again, this is correct since any value change of $v$ can be its last value change, and, in particular, $v$ may remain unchanged in a plan for a given problem.

Finally, the POP-PCG algorithm starts with a null plan that contains the end operator $A_i^*$ only if $v_i^*$ is specified by $Goal$.

## 4. Directed-Path Singly Connected and General DAGs

In this section we analyze planning complexity in face of more complicated causal graphs. First, we show that when the causal graph is directed-path singly connected even plan existence is np-complete. Second, we show that for general causal graphs the situation is even worse. Finally, we characterize an important parameter of the causal graph affecting planning complexity, which allows us to extend the class of problems which are in np.

**Theorem 2** *Plan existence for* strips *planning problems with unary operators and directed-path singly connected causal graph is* np-*complete.*

**Proof:**  For the proof see Appendix B, p. 346.  □

Note that node indegree in the causal graph of the problem created in the proof of Theorem 2 is bounded by 6. The hardness for directed-path singly connected causal graphs with maximal indegree lower than 6 is thus open.

The directed-path singly connected structure of the causal graph turns out to be crucial for guaranteeing reasonable solution times. As we now show, there are solvable propositional planning problems with an arbitrary acyclic (DAG) causal graph that have minimal solutions of exponential size. Analysis of this class of problems points to the reason for such provable intractability. This allows us to characterize an important parameter of the causal graph affecting planning complexity and to extend the class of problems which are in np. However, all these restricted problems are still np-complete.

**Theorem 3** *Plan generation for general* strips *planning problems with unary operators and acyclic causal graph is provably intractable, i.e. it is harder than* np.

This theorem follows from Theorem 5.4 in (Jonsson & Bäckström, 1998b), that shows that plan generation for the 3S problem class is provably intractable. The point is that the upper bound for MinPlanSize, presented in Eq. 5, can be exponential in the size of the input in this case. First, we show by example that this upper bound can be achieved, then we present some analysis of the reasons for this intractability.

The following example shows that an exponential upper bound can be achieved. It was used in the proof of Theorem 5.4 in (Jonsson & Bäckström, 1998b), and was originally presented in a different context by Bäckström and Nebel (1995). Consider a propositional planning problem with $|\mathcal{V}| = n$, where, for $1 \leq i \leq n$, $\mathcal{D}(v_i) = \{0,1\}$ and $\mathsf{pred}(v_i) = \{v_1, \ldots v_{i-1}\}$. The operator set $\Lambda$ consist of $2n$ operators $\{A_1, A_1', \ldots A_n, A_n'\}$ where

$$\mathsf{pre}(A_i)[j] = \mathsf{post}(A_i')[j] = \left\{ \begin{array}{ll} 0 & \text{if } j = i \\ \mathsf{u} & \text{otherwise} \end{array} \right.$$

$$\mathsf{pre}(A_i')[j] = \mathsf{post}(A_i)[j] = \left\{ \begin{array}{ll} 1 & \text{if } j = i \\ \mathsf{u} & \text{otherwise} \end{array} \right.$$

$$\mathsf{prv}(A_i)[j] = \mathsf{prv}(A_i')[j] = \left\{ \begin{array}{ll} 0 & \text{if } j < i-1 \\ 1 & \text{if } j = i-1 \\ \mathsf{u} & \text{otherwise} \end{array} \right.$$

It is easy to see that the causal graph of this problem forms a DAG (see Figure 7), and an instance of this planning problem with the initial state $\langle 0, \ldots, 0 \rangle$ and the goal state $\langle 0, \ldots, 0, 1 \rangle$ has a unique minimal solution of length $2^n - 1$ corresponding to a Hamilton path in the state space.

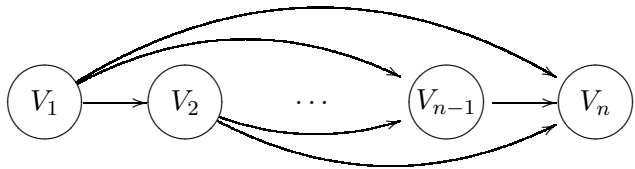

Figure 7: Causal graph for the proof of Theorem 3

Now we show that this escalation in complexity can be "parametrized" by the form of the causal graph.

**Lemma 3** *For any solvable problem instance $\Pi$ with an acyclic causal graph over $n$ variables, for any variable $v$, we have that:*

$$\mathsf{MaxReq}(v_i) \leq 1 \ + \ \sum_{j=i+1}^{n} \rho(v_i, v_j)$$

*where $\rho(v_i, v_j)$ denotes the total number of different, not necessary disjoint, paths from $v_i$ to $v_j$, where variables are ordered via a topological sort of the causal graph.*

**Proof:** The proof is by induction on $i$. For $i = n$ it is obvious that $\mathsf{MaxReq}(v_n) \leq 1$. Now we assume that the lemma holds for any $i > k$, and prove it for $i = k$. Without loss of generality, assume that $\mathsf{succ}(v_k) \neq \emptyset$. Otherwise, we simply have that $\mathsf{MaxReq}(v_k) \leq 1$.

The proof is straightforward:

$$\mathsf{MaxReq}(v_k) \overset{Eq.\ 1}{\leq} 1 \ + \sum_{v_{i_k} \in succ(v_k)} \mathsf{MaxReq}(v_{i_k}) \leq$$

$$\overset{I.H.}{\leq} 1 + |\mathsf{succ}(v_k)| + \sum_{v_{i_k} \in succ(v_k)} \sum_{j=i_k+1}^{n} \rho(v_{i_k}, v_j) =$$

$$= 1 + \sum_{j=k+1}^{n} \rho(v_k, v_j)$$

$\square$

Lemma 3 entails that the upper bound for $\mathsf{MinPlanSize}(\Pi)$ for a general planning problem with unary operators and acyclic causal graph depends on the number of different paths between the nodes in the causal graph. An immediate conclusion is that there is a significant class of problems with an acyclic causal graph for which planning is in NP. Let a DAG be called *max-$\delta$-connected* if the number of different directed paths between every two nodes in this graph is bounded by $\delta$.

**Theorem 4** *Plan generation for* STRIPS *planning problems with unary operators and max-δ-connected causal graph is* NP-*complete if δ is polynomially bounded.*

**Proof:** Membership in NP is straightforward: If the variables of a given problem Π are considered in a topological ordering induced by the causal graph, then from Lemma 3 follows that, for any variable $v_i$, $\mathsf{MaxReq}(v_i) \leq \delta n$. In turn, from this follows that $\mathsf{MinPlanSize}(\Pi) \leq \delta n^2$, and thus, if δ is polynomially bounded, then we can guess a minimal plan for Π that could be verified in polynomial time.

The hardness follows from Theorem 2 that shows that even if the causal graph is max-1-connected (directed-path singly connected), then plan existence (and thus plan generation) is hard. $\square$

## 5. Serializable Subgoals

A set of subgoals is defined to be *serializable* (Korf, 1987) if there exists an ordering among the subgoals such that the subgoals can always be solved sequentially without ever violating a previously solved subgoal in the order. Naturally, not all collections of subgoals are serializable – sometimes it may be necessary to interleave plans for achieving different goals. However, when a problem instance is *serially decomposable*, it is possible to design a set of macro-operators with respect to which the subgoals *are* serializable (Korf, 1985).

A problem instance is serially decomposable if there exists some ordering of the state variables for which the effect of each operator on each state variable depends only on that state variable and previous state variables in the ordering. Unfortunately, Bylander (1992) shows that determining serial decomposability of a problem is PSPACE-complete.

One major open problem put forth by Bylander in this context is: *If a problem is known to be serially decomposable, how difficult is it to determine whether a given instance is solvable?* As far as we know, the only work in this direction was done by Chalasani *et al.* (1991), where the serial decomposability of the "general permutation problem" was considered. In particular, they showed that this problem is in NP, but it is unknown whether it is NP-hard. Recently, some complementary results for Bylander's question were presented Köehler and Hoffmann (2000). Our results shed more light on this question: Any problem instance based on a unary operator domain whose causal graph is acyclic is serially decomposable. Therefore, it can be concluded that finding a solution for serially decomposable problems may require exponential time (i.e., the problem is in EXPTIME). However, Bylander's question is about plan existence. In that case, Theorem 3 does not apply, and we can only apply our NP-hardness result (for directed-path singly-connected graphs), since it addresses plan existence as well.

Weld (1999) hypothesized that: (1) If the underlying causal graph of the planning problem is acyclic, then a serialization ordering on the subgoals of the problem is obvious; (2) *Serialized* subgoals could be solved extremely quickly because no backtracking is required between them. Although the first observation sounds intuitive, our results suggest that it is rarely true. The acyclicity of the causal graph implies serializability, but in most of the cases its structure does not provide us sufficient information about the actual serialization ordering. Even when the causal graph is a directed tree one must think first before choosing an ordering. Likewise, our results imply that when the causal graph does not form an undirected tree determining a subgoal ordering is NP-complete, and if the causal graph is not directed-path singly connected, the problem is even more complex.

The second observation is not always true either. The problem is that it is important to determine not only the serialization ordering over the subgoals, but also the exact strategies for achieving them. As we showed, in certain cases, a problem with $n$ serializable subgoals requires an exponentially long solution. When the domain variables are not binary, the situation is even worse – some of the corresponding complexity results can be derived from the computational analysis of Domshlak and Dinitz (2001).

## 6. Connection with Related Work on Planning Complexity

The idea of analyzing and exploiting structural properties is not new to classical planning, and in the last few years a number of important results have emerged. Generating plans in the context of the STRIPS representation language was shown by Bylander (1994) to be PSPACE-complete. Despite this fact, the existence of many successful planning systems, especially in recent years, demonstrates that planning is possible and practical for a wide list of domains. Bylander argues that the large gap between the theoretical hardness of planning and its practical success stems from the use of domain-dependent problem analysis and algorithms. Consequently, various authors have explored the existence of some constrained problem classes for which planning is easier.

In this section we shortly overview some of the major, previous results on complexity of planning, and discuss their relationship to the results presented in this paper. For a more detailed presentation of the previous results discussed below we refer the reader to the original papers.

### 6.1 Local Syntactical Restrictions

In his seminal paper, Bylander (1994) presents a number of complexity results for propositional planning, analyzing different planning problems based on the type of formulas used, the number and type (positive/negative) of operator pre- and postconditions, etc. The work of Bylander is extended by some interesting, complementary results by Erol *at al.* (1995). For example, Bylander shows that propositional planning in domains where each operator is restricted to have positive preconditions and one postcondition only is tractable. Generally, extremely severe restrictions on operators are required to guarantee tractability, or even membership in NP. Note that Bylander (1994) and Erol *et al.* (1995) focuses on *local syntactical* properties of operators, i.e., properties of single operators.

The only syntactic restriction that we pose on the planning problems in this paper is the unarity of the operators. Determining plan existence for this, apparently easier class of problems was shown by Bylander to be as hard as general propositional planning, i.e. PSPACE-complete. Note that this result by itself does not entail our Theorem 3, since planning problems with unary operators may induce causal graphs with cycles. Therefore, none of our results is entailed by the results presented by Bylander (1994) and Erol *et al.* (1995).

## 6.2 Global Syntactical Restrictions

Bäckström and Klein (1991a, 1991b), and, subsequently, Bäckström and Nebel (1995), consider other types of restrictions, but using a more refined model (the *SAS* formalism) in which:

1. The state variables are multi-valued, and

2. Two types of preconditions are considered: *prevail* conditions, which are variable values that are required prior to the execution of the operator and are not affected by the operator, and *preconditions*, which are affected by the operator.

In general, four different restrictions were considered in these works:

(P) *Post-uniqueness*: For each effect there is at most one operator that achieves this effect. In other words, desired effects determine operators to be used in a plan. Formally, a problem instance is post-unique if and only if, for each $v_i \in \mathcal{V}$ and $x \in \mathcal{D}(v_i)$, there is at most one operator $A \in \Lambda$ such that $\mathsf{post}(A)[i] = x$.

(S) *Single-valuedness*: At most one value of each state variable appears in the prevail conditions of the operators. For instance, if a certain operator requires the light to be on (as a prevail condition), no other operator can use the prevail condition that the light is off. Formally, a problem instance is single-valued iff there exist no two operators $A, A' \in \Lambda$ and $v_i \in \mathcal{V}$ such that $\mathsf{prv}(A)[i] \neq \mathsf{u}$, $\mathsf{prv}(A')[i] \neq \mathsf{u}$, and $\mathsf{prv}(A)[i] \neq \mathsf{prv}(A')[i]$.

(U) *Unariness*: Each operator affects only one state variable.

(B) *Binariness*: All state variables have exactly two possible values, i.e. all state variables are propositional.

All these four properties are syntactical. However, the properties P and S differ from the properties U and B by the fact that they have a *global* nature: Post-uniqueness and single-valuedness restrict not the form of the operators, but a global property of the whole set of operators. Bäckström and Nebel (1995) showed that US (unariness and single-valuedness) is the extreme problem class for which plan generation is polynomial. [6]

The problems that we analyzed in this paper belong to the problem class UB, by definition. As we already mentioned, even determining plan existence for this class of problems is PSPACE-complete. Now consider the problem class PUB. Bäckström and Nebel (1995) showed that: (i) PUB has instances with exponentially long minimal solutions, thus plan generation for PUB is requires exponential time; (ii) existence of bounded length plans for PUB is strongly NP-hard; and (iii) the complexity of general plan existence for PUB is still an open question. Informally it means that strengthening restrictions from UB to PUB does not reduce the complexity significantly, at least from the practical point of view.

**Proposition 1** *Every UB problem instance with a tree causal graph is either post-unique, or can be transformed into an equivalent post-unique problem instance in (low) polynomial time. Thus, TreeUB $\subset$ PUB.*

---

6. For a thorough analysis of the complexity of SAS planning, we refer to Bäckström and Nebel (1995).

**Proof:** Consider a UB problem with a tree causal graph, and suppose that it is not post-unique. It means that there exist a variable $v \in \mathcal{V}$, with $\mathcal{D}(v) = \{v', v''\}$, such that there exist two operators $A_1, A_2 \in \Lambda_v$ that change the value of $v$ from $v'$ to $v''$, and $\mathsf{prv}(A_1) \neq \mathsf{prv}(A_2)$.

From the assumption that the causal graph forms a tree it follows that $|\mathsf{pred}(v)| \leq 1$. If $\mathsf{pred}(v) = \emptyset$, then it is easy to see that the existence of such a pair of operators is simply impossible. Therefore, let $\mathsf{pred}(v) = \{w\}$, where $\mathcal{D}(w) = \{w', w''\}$. Without loss of generality assume that $\mathsf{prv}(A_1)[w] = \{w'\}$, $\mathsf{prv}(A_2)[w] = \{w''\}$. Otherwise, if, for instance, $\mathsf{prv}(A_1)[w] = \mathsf{u}$, then it is easy to see that $A_2$ is a redundant operator.

Observe that in this case, prevail dependence of $v$ on $w$ is redundant: We can replace the pair of operators $A_1, A_2$ in $\Lambda$ by a single operator $A$ that changes the value of $v$ from $v'$ to $v''$ without any prevail condition. The replacement of $A_1, A_2$ by $A$ brings us to an *equivalent* problem instance in which the operator set $\Lambda_v$ is post-unique. This way we continue to process iteratively all such "problematic" variables $v$ until we arrive at a post-unique problem instance. $\square$

**Proposition 2** *There are UB problem instances with a tree causal graph that are not single-valued, thus TreeUB $\not\subseteq$ UBS.*

**Proof:** The proof of Proposition 2 is straightforward: Consider a variable $v \in \mathcal{V}$, $\mathcal{D}(v) = \{v', v''\}$, such that $\mathsf{succ}(v) = \{u, w\}$. It can be the case that any value change of $u$ will be prevailed by $v'$, while any value change of $w$ will be prevailed by $v''$. Therefore, restricting causal graphs even to trees does not entail single-valuedness.[7] $\square$

Propositions 1 and 2 show that TreeUB is a polynomial subclass of PUB that is not entailed by any tractability results of Bäckström and Nebel (1995).

**Proposition 3** *There are UB problem instances with a polytree causal graph that are neither single-valued, nor post-unique.*

**Proof:** The proof is straightforward: Consider a planning problem with a polytree causal graph, such that there exist a variable $v \in \mathcal{V}$ with $\mathsf{pred}(v) = \{u, w\}$, and the following operator set $\Lambda_v$:

| pre | post | prv |
|-----|------|-----------|
| $v'$ | $v''$ | $\{u', w''\}$ |
| $v'$ | $v''$ | $\{u'', w'\}$ |
| $v''$ | $v'$ | $\{u', w'\}$ |
| $v''$ | $v'$ | $\{u'', w''\}$ |

Clearly, any problem instance with such $\Lambda_v \subseteq \Lambda$ is neither single-valued, nor post-unique, since (i) there is more than one operator achieving any value of $v$, and (ii) both values of $u$ (and both values of $w$) appear in prevail conditions of the operators in $\Lambda_v$. Note that the maximal indegree of such a polytree can be minimal, i.e. equal to 2. Thus, the proposition is valid for any polytree that is not a tree. $\square$

---

7. Using the simple construction technique from the proof of Proposition 1 it can be shown that restricting causal graphs to directed chains only does entails single-valuedness. However, this case is too restrictive.

From Proposition 3 it follows that Theorems 1 and 4 introduce new polynomial and NP-easy subclasses of the UB problem class, respectively.

## 6.3 Structural Restrictions in Propositional Planning

Jonsson and Bäckström (1998b) present the 3S class of planning problems. This class is most closely related to the problems examined in this paper, since it defines a special subclass of problems with binary variables, unary operators and acyclic causal graphs. The 3S problem class is defined by posing some additional, relatively severe, restrictions on the problem's operator set: Each variable $v$ in a 3S problem instance is required to be either (i) *static*, i.e., unchangeable; (ii) *symmetrically reversible*, i.e., for each operator $A$ affecting $v$, there exist an operator $A'$ affecting $v$ with the same prevail conditions and the opposite effect; or (iii) *splitting*. For the formal definition of the splitting property we refer to Jonsson and Bäckström (1998b). Informally, if a binary variable $v$ is splitting then the problem instance can be split into three, well-defined subproblems that can be solved independently. For this class of planning problems it was shown that plan existence can be determined in polynomial time, while plan generation is provably intractable, since there are instances of 3S with exponentially long minimal solutions. In particular, the problem instance that we used in the proof of Theorem 3 is in 3S.

The complexity analysis by Jnonsson and Bäckström (1998b) is somewhat unique in the research on complexity of propositional planning, since, to the best of our knowledge, this was the only attempt to exploit not only syntactical restrictions on the operator set, but also some structural restrictions on interaction between the variables. Our analysis can be seen as continuing this direction by looking on the structural restrictions *only*. We believe that eliminating the marginal effect of the problem structure on the problem's (potential) hardness will allow us to understand better the connection between the component interactions topology, and the potential complexity of the problem.

## 6.4 Structural Restrictions in Multi-valued Formalisms

When the variables are no longer propositional, some additional properties of the problems can be identified, and, possibly, exploited. In particular, additional internal structures of the problem can be analysed.

Jonsson and Bäckström (1998a) analyze different properties of a multi-valued problem structure, which is called the *domain transition graph*. Such a structure is defined for each state variable of the problem, and it describes possible transitions between different values of this variable. The domain transition graph of a state variable $v$ is a directed labeled graph $G_v = (V, E)$, where $V$ is associated with the $v$'s set of possible values, $\mathcal{D}(v)$, and $(x, A, y) \in E$ if and only if the operator $A$ can be applied at some state in which $v = x$, and its application results in a state in which $v = y$ holds.

Jonsson and Bäckström identify sets of structural restrictions on domain transition graphs which make planning instances tractable. Roughly, the properties are the following: (1) The problem domain is *interference-safe*, i.e., each operator is either unary or irreplace-able with respect to every variable it affects. An operator $A$ is *irreplaceable* with respect to a variable $v$ if the removal of all edges from $G_v$ that stem from $A$ disconnects some weakly connected component of $G_v$. (2) For every variable $v$, the graph $G_v$, restricted to the set of values that appear in the prevail conditions of some operators, is acyclic. (3) Any sequence of operators annotating a path from $x$ to $y$ in the domain transition graph of $v$, is stronger than all shortest such sequences connecting $x$ and $y$. Here, a sequence $A_1, \ldots, A_k$ is *stronger* than $A'_1, \ldots, A'_l$ if there is a subsequence $A_{i_1}, \ldots, A_{i_l}$ of $A_1, \ldots, A_k$ such that for every $1 \leq j \leq l$, the prevail conditions of $A'_j$ are a subset of the prevail conditions of $A_{i_j}$. Jonsson and Bäckström present a map of the computational complexity of problems with different restrictions, displaying the frontier between the tractable and intractable cases.

Each domain transition graph combines and structures the influence of many operators on a particular variable. Therefore, they provide us a more global picture than the operator set alone. Hence, in spite of the fact that domain transition graphs do not capture the relationship between different variables, they do allow us to express some structural properties that address interactions between the variables (e.g., see property (2) above).

Observe that domain transition graphs are not very informative in the case of propositional planning, since they are only distinguish between the variables that can be changed only in one direction and the variables that can be changed in both directions. Although this property of domain transition graphs allows to distinguish between the polynomial planning with only positive postconditions, and the PSPACE-complete planning with both positive and negative postconditions (Bylander, 1994), it seems to be not very helpful in further hierarchical refinement of the propositional planning complexity. On the other hand, there is no a priori reason why the causal graphs will not be informative in the multi-valued case. Exploiting the properties of causal graphs, together with the properties of domain transition graphs, seems to be a natural direction to extend the work presented in this paper. The recent work of Domshlak and Dinitz (2001) on multi-entity off-line coordination can be seen as investigating connections between the structure of the causal graph, together with the properties of the domain transition graphs, and the complexity of the corresponding problems in case of multi-valued domains. To the best of our knowledge, this is the only work that was done with respect to such a "mixed" structural analysis, and a lot of work remains to be done. For instance, combining various properties of the domain transition graphs studied by Jonsson and Bäckström (1998a), with the properties of the problem's causal graph is a direction for the further research.

## 7. Summary and Future Work

We have shown that the form of the causal graph for STRIPS planning problems with unary operators is an important factor in determining the computational complexity of plan generation. In particular, we have shown that a polynomial time algorithm exists for any problem with a polytree causal graph and the node indegree bounded by a constant. More generally, this result shows that planning with polytree causal graphs is at most (what is often referred to in the Bayes nets literature as) locally exponential, i.e., it is exponential in the maximal number of parents of a node. Note that in hardware-control planning problems the maximal node indegree is expected to be small, since prevail dependencies between the variables reflect the direct interconnections between the corresponding hard-ware components. Likewise we have shown that for a problem with directed-path singly connected causal graph the maximal plan length is a low order polynomial, but the problem is NP-complete. More generally, we have shown a relation between the number of paths between variables in the causal graph and the computational complexity of the corresponding planning problem. Finally we have presented the impact of our results on the question of complexity of planning problems with serializable subgoals, and connected our work with previous results on planning complexity.

Our work leaves a number of open questions with respect to purely syntactical, and a mixture of both structural and syntactical restrictions on the planning problems with unary operators. In the former case, one of the most important directions is a further analysis of causal graphs with constantly bounded node indegree. It turns out that complexity analysis for this class of problems will be very helpful in understanding various computational properties of CP-nets (Boutilier et al., 1999). Although here we provided a partial answer for this question, the general picture of the worst-case complexity for this class of problems is not clear. For example, if the indegree of the causal graph is known to be bounded by 2, and this is the only structural property of the causal graph, it is even not clear whether this problem subclass is in NP.

In the latter case, various syntactical restrictions can be analysed together with the form of the causal graph. For example, one may be interested in the computational properties of the problems with acyclic causal graphs, and the restriction that every operator has at most $\alpha$ prevail conditions, where $\alpha$ is bounded by a constant. This, as well as many other related questions with respect to various special cases of planning with unary operators are of interest for the future work.

## Acknowledgments

A preliminary version of this paper appeared in the Sixth International Conference on Artificial Intelligence Planning and Scheduling, April, 2002. We would like to thank the three anonymous reviewers for their extremely helpful comments. Ronen Brafman is supported in part by the Paul Ivanier Center for Robotics Research and Production Management.

## Appendix A. A Short Review of POP, Causal Links and Threats

We represent a plan as a tuple: $\langle \mathcal{A}, \mathcal{O}, \mathcal{L} \rangle$, where $\mathcal{A}$ is a set of unary *operators*, $\mathcal{O}$ is a set of *ordering constraints* over $\mathcal{A}$, and $\mathcal{L}$ is a set of *causal links*. For example, if $\mathcal{A} = \{A_1, A_2, A_3\}$ then $\mathcal{O}$ might be the set $\{A_1 < A_3, A_2 < A_3\}$. These constraints specify a plan in which $A_3$ is necessarily the last operator, but do not commit to a particular order on $A_1$ and $A_2$. Naturally, the set of ordering constraints must be consistent, i.e., there must exist some total order satisfying them. A causal link has the form $A_p \xrightarrow{\vartheta_i} A_c$, where $A_p$ and $A_c$ are operators and $\vartheta_i$ is a possible value for some propositional variable $v_i$. It denotes the fact that $A_p$ produces (i.e., has the postcondition) $v_i = \vartheta_i$ which is consumed by $A_c$ (i.e., used to satisfy a pre- or prevail-condition of $A_c$). Causal links help us detect whether one operator $A_t$ interferes with the work done to enable the execution of some other operator $A_c$. In that case, $A_t$ is said to constitute a *threat* to one of $A_c's$ causal links. Formally, suppose that $\langle \mathcal{A}, \mathcal{O}, \mathcal{L} \rangle$ is a plan, and $A_p \xrightarrow{\vartheta_i} A_c$ is a causal link in $\mathcal{L}$. Let $A_t$ be a different operator in $\mathcal{A}$. We say that $A_t$ *threatens* $A_p \xrightarrow{\vartheta_i} A_c$ when the following two criteria are satisfied:

- $\mathcal{O} \cup \{A_p < A_t < A_c\}$ is consistent, and

- $A_t$ has $\neg \vartheta_i$ as an effect.

When a partial order plan $P$ contains threats, it is possible that the goal will not be achieved by some (or all) of the total order plans consistent with $P$'s ordering constraints. To prevent this, the plan generator must check for threats and remove them by adding one of two simple ordering constraints: $A_t < A_p$ (demotion) or $A_c < A_t$ (promotion).

A tutorial introduction to POP algorithms can be found in (Weld, 1994). POP is a regressive framework for partial order planning that starts with the null plan and continuously updates it by inserting new actions and removing threats. This process continues until the precondition and the prevail conditions of every operator in the plan are supported by some causal link and no threats exist. The first argument to POP is a plan and the second argument is an agenda of goals that need to be supported by causal links. Each item on the agenda is represented by a pair $\langle \vartheta_i, A \rangle$ where $\vartheta_i$ is either pre- or prevail condition of a plan action $A$. The last argument to POP is the whole collection of the operators defined by the planning instance. The initial call to POP contains the null plan, a specially initialized agenda, and the operator set $\Lambda$ of the given problem.

In this paper we introduce a specialized, *deterministic* POP algorithm that starts the planning process using a variant of the null plan which encodes the planning problem. In particular, if the planning instance has $v_1^*, \ldots, v_n^*$ as the goal then the corresponding null plan has exactly $2n$ dummy unary operators, $\mathcal{A} = \{A_1^0, \ldots, A_n^0, A_1^*, \ldots, A_n^*\}$, $n$ ordering constraints, $\mathcal{O} = \{\{A_1^0 < A_1^*\}, \ldots, \{A_n^0 < A_n^*\}\}$, and no causal links, $\mathcal{L} = \{\}$. For every $v_i \in \mathcal{V}$, $A_i^0$ is the corresponding "start" operator - it has neither pre- nor prevail conditions, and its effect specifies the value of the variable $v_i$ in the initial state, which is denoted by $v_i^0$. Similarly, $A_i^*$ is the "end" operator - it has no effect, no prevail conditions, but its precondition is set to the value of $v_i$ in the goal state, which in turn is denoted by $v_i^*$.[8]

---

8. Actually, the goal state may not specify the values of all the variables, thus the number of the end operators can be less than $n$. However, for clarity of presentation, we leave this definition of the null plan.

Our description of the null plan is modified from that of Weld (1994) to better suit the restriction to unary operators. Likewise, the initial call to our POP algorithm contains the agenda $\{\langle v_1^*, A_1^* \rangle, \ldots, \langle v_n^*, A_n^* \rangle\}$.

## Appendix B. Proofs and Auxiliary Results

**Lemma 2** *If* FORWARD-CHECK *was successful then* POP-PCG *will return a valid plan.*

**Proof:** The lemma will follow from the following claims:

1. For every agenda item, there exists an operator that has it as an effect.

2. There are no threats in the output of POP-PCG.

3. The ordering constraints in $\mathcal{O}$ are consistent.

4. The agenda will be empty after a polynomial number of steps.

(1+4) The first claim follows from the success of the FORWARD-CHECK procedure. FORWARD-CHECK implies that for any $\nu_i^j \in \sigma(v_i)$ there is an operator instance $A(\nu_i^j) \in \Gamma_i \cup \{A_i^0\}$. Therefore, if $\nu_i^j \in \sigma(v_i)$ then the existence of an appropriate $A_{add}$ is promised.

Assume to the contrary that $\nu_i^j \notin \sigma(v_i)$ and, without loss of generality, assume that this is the first iteration that it happens. If so, then for each variable $u \in \mathsf{succ}(v_i)$, there is no edge labeled by $\nu_i^j$ in the graph $G'(u)$, which is created by the FORWARD-CHECK. From this follows that $A_{need}$ cannot have $\nu_i^j$ as a prevail condition, and thus $A_{need}$ has to affect the variable $v_i$ itself. In this case either $A_{need} = A(\nu_i^{j+1})$ or $A_{need} = A_i^*$.

Consider the former case: If $A_{need} = A(\nu_i^{j+1})$ then $\nu_i^{j+1}$ was previously selected from the agenda. By our assumption it means that $\nu_i^{j+1} \in \sigma(v_i)$, and it contradicts our assumption that $\nu_i^j \notin \sigma(v_i)$ since $\nu_i^j$ is a predecessor of $\nu_i^{j+1}$ on $\sigma(v_i)$.

Now consider the last option that $A_{need} = A_i^*$. If $A_{add} = A(\nu_i^j)$ then the goal value of the variable $v_i$ is consistent with $\nu_i^j$, and $A(\nu_i^{j-1}) \in \mathcal{A}$ (see Step 3(b)ii). If $A(\nu_i^{j-1}) \in \mathcal{A}$ then $\nu_i^{j-1}$ was previously selected from the agenda. By our assumption it means that $\nu_i^{j-1} \in \sigma(v_i)$. However, it contradicts our assumption that $\nu_i^j \notin \sigma(v_i)$ since $\sigma(v_i)$, by definition, terminates with a node consistent with the goal value of $v_i$.

In addition, since we have shown that the only operators added into $\mathcal{A}$ are those from $\Gamma_i \cup \{A_i^0, A_i^*\}$, for $1 \leq i \leq n$, the agenda will be empty after $O(n^2)$ steps.

(2) Suppose that some operator $A_t$ threatens $A_p \xrightarrow{\vartheta_i} A_c$, i.e.,

- $\mathcal{O} \cup \{A_p < A_t < A_c\}$ is consistent, and

- $A_t$ has $\neg \vartheta_i$ as an effect.

For a given variable $v_i$, POP-PCG forces the operators affecting $v_i$ as follows (Step 4):

$$A_i^0 \equiv A(\nu_i^1) < A(\nu_i^2) < \ldots < A(\nu_i^x), \quad x \leq \mathsf{FMaxReq}(v_i) \tag{4}$$

Thus $A_c$ can only be an operator with $\vartheta_i$ as a prevail condition. Note that $A_p$ and $A_t$ affect the same variable $v_i$. In (1) we already showed that $\vartheta_i = \nu_i^j \in \sigma(v_i)$. In that case $A_t = A(\nu_i^l)$, $l > j$. However, if $\nu_i^j$ is a prevail condition of $A_c$ then the ordering constraint $\{A_c < A(\nu_i^{j+1})\}$ was added to $\mathcal{O}$ at Step 6. If so then from Eq. 4, it follows that $\{A(\nu_i^l) < A(\nu_i^{j+1}), A(\nu_i^{j+1}) < A(\nu_i^l)\}$, $l > j$, will be implied by $\mathcal{O} \cup \{A_p < A_t < A_c\}$, and it contradicts with the assumption that $\mathcal{O} \cup \{A_p < A_t < A_c\}$ is consistent.

(3) The ordering constraints are consistent if no two operators $A_i$ and $A_j$ are such that $\mathcal{O}$ implies $\{\{A_i < A_j\}, \{A_j < A_i\}\}$. In what follows, $A_i$ will be used to denote an arbitrary operator affecting variable $v_i$.

First note that each ordering constraint added in Step 4 or Step 6 is either between operators affecting the same variable or between operators affecting a variable and its child (with respect to the causal graph). In particular, if $A_i < A_j$ was added in Step 4 then either $v_i = v_j$ or $v_i \in \mathsf{pred}(v_j)$, whereas if $A_i < A_j$ was added in Step 6 then $v_j \in \mathsf{pred}(v_i)$.

Assume, to the contrary that $\mathcal{O}$ implies $A_i < A_j$ and $A_j < A_i$. From the argument above, we know that there is a, possibly empty, path between $v_i$ and $v_j$ in the undirected graph induced by the causal graph. By our structural assumption, we know that such an undirected path between $v_i$ and $v_j$ is unique, and thus the situation is as follows: We have two chains of operators

$$\alpha \ : \ A_i = A_{i_0}^1 < \ldots < A_{i_0}^{x_0} < A_{i_1}^1 < \ \ldots \ < A_{i_1}^{x_1} < \ldots \ldots < A_{i_m}^1 < \ldots < A_{i_m}^{x_m} = A_j$$
$$\beta \ : \ A_i = \bar{A}_{i_0}^1 > \ldots > \bar{A}_{i_0}^{y_0} > \bar{A}_{i_1}^1 > \ \ldots \ > \bar{A}_{i_1}^{y_1} > \ldots \ldots > \bar{A}_{i_m}^1 > \ldots > \bar{A}_{i_m}^{y_m} = A_j$$

such that, for $0 \leq k \leq m$, both $x_k \geq 1$ and $y_k \geq 1$. The corresponding unique undirected path between $v_i$ and $v_j$ is:

$$v_i = v_{i_0} \cdot v_{i_1} \cdot \ldots \cdot v_{i_{m-1}} \cdot v_{i_m} = v_j$$

Without loss of generality, the internal elements of $\alpha$ and $\beta$ are disjoint. Otherwise, if there is an operator $B$ that belongs to the internal parts of both $\alpha$ and $\beta$ then we can reduce these chains and deduce $A_i < B$ and $A_i > B$.

The proof of consistency is as follows:

(a) We prove that if such $\alpha$ and $\beta$ exist then at least one of them should have at least one internal element.

(b) We show some useful property of $\alpha$ and $\beta$, that is exploited in (c).

(c) We show that for $0 \leq k \leq m$, $A_{i_k}^{x_k}$ and $\bar{A}_{i_k}^{y_k}$ are different except when $x_0 = y_0 = 1$.

   Note that (a) together with (c) contradicts our assumption that $A_{i_m}^{x_m} = \bar{A}_{i_m}^{y_m}$.

(a) Assume, to the contrary, that both $\alpha$ and $\beta$ do not contain any internal elements. If so, then the algorithm actually adds to $\mathcal{O}$ ordering constraints $A_i < A_j$ and $A_j < A_i$. If $v_i$ and $v_j$ are the same variable then $A_i < A_j$ can only stem from Step 4 and only because $A_i$ has the precondition of $A_j$ as its effect. However, by definition of FORWARD-CHECK, $A_j$ can not have the same role w.r.t. $A_i$ and thus it is impossible that $A_j < A_i$ was added to $\mathcal{O}$. Alternatively, if $v_i$ is a parent of $v_j$ then $A_i < A_j$ can stem only from Step 4 because $A_i$ has the prevail condition of $A_j$ as its effect. Suppose that $A_i = A(\mathsf{b}_i^j)$ and thus $\mathsf{b}_i^j \in \mathsf{prv}(A_j)$. In turn, $A_j < A_i$ can be added only in Step 6 because $A_j$ has the precondition of $A_i$ as the prevail condition. But if $\mathsf{b}_i^j \in \mathsf{prv}(A_j)$ then $A_i = A(\mathsf{w}_i^j)$, and this contradicts our assumption that $A_i = A(\mathsf{b}_i^j)$. Alternatively, we can assume that $A_i = A(\mathsf{w}_i^j)$ but this situation is completely symmetric, and thus the result will be the same. Hence we proved that either $\alpha$ or $\beta$ have to contain at least one internal element. In particular it means that the next to last elements of $\alpha$ and $\beta$ are different and this fact is exploited later in the proof.

(b) Consider subchains of $\alpha$ that consist of operators affecting only one particular variable. For each such a subchain, i.e. for $0 \leq k \leq m$, for $1 \leq j \leq x_k - 1$, the ordering constraint $A_{i_k}^j < A_{i_k}^{j+1}$ can only stem from Step 4 because $A_{i_k}^j$ has the precondition of $A_{i_k}^{j+1}$ as its effect. Thus, $\mathsf{post}(A_{i_k}^j) = \mathsf{pre}(A_{i_k}^{j+1})$. Similarly, for the subchains of $\beta$, for $0 \leq k \leq m$ and for $2 \leq j \leq y_k$, $\mathsf{post}(\bar{A}_{i_k}^j) = \mathsf{pre}(\bar{A}_{i_k}^{j-1})$. In what follows we denote this property of $\alpha$ and $\beta$ by *local monotonicity*.

(c) First suppose that either $x_0 > 1$ or $y_0 > 1$, or both. Consider the following sequence:

$$\gamma \ : \ \bar{A}_{i_0}^{y_0} < \bar{A}_{i_0}^{y_0-1} < \ldots < \bar{A}_{i_0}^1 = A_{i_0}^1 < \ldots < A_{i_0}^{x_0}$$

From the local monotonicity, the construction of FORWARD-CHECK, and the fact that $|\gamma| \geq 2$ it follows that $\mathsf{post}(\bar{A}_{i_0}^{y_0})$ appears before $\mathsf{post}(A_{i_0}^{x_0})$ on the maximal sequence $\sigma_{v_{i_0}}$. Continuing with the next variable $v_{i_1}$ we claim that $\mathsf{post}(\bar{A}_{i_1}^{y_1})$ has to appear before $\mathsf{post}(A_{i_1}^{x_1})$ on $\sigma_{v_{i_1}}$.

*(i)* If $v_{i_0}$ is a parent of $v_{i_1}$ then $A_{i_0}^{x_0} < A_{i_1}^1$ can only stem from Step 4 because $A_{i_0}^{x_0}$ has the prevail condition of $A_{i_1}^1$ as its effect. In turn, $\bar{A}_{i_0}^{y_0} > \bar{A}_{i_1}^1$ can only stem from Step 6 because the prevail condition of $\bar{A}_{i_1}^1$ is the precondition of $\bar{A}_{i_0}^{y_0}$. From the relation between $A_{i_0}^{x_0}$ and $\bar{A}_{i_0}^{y_0}$, and the construction of $G'_e(v_{i_1})$ in FORWARD-CHECK, it follows that $\mathsf{post}(\bar{A}_{i_1}^1)$ appears before $\mathsf{post}(A_{i_1}^1)$ on $\sigma(v_{i_1})$. Subsequently, from the local monotonicity it follows that $\mathsf{post}(\bar{A}_{i_1}^{y_1})$ appears before $\mathsf{post}(A_{i_1}^{x_1})$ on $\sigma(v_{i_1})$.

*(ii)* Similarly, if $v_{i_1}$ is a parent of $v_{i_0}$ then $\bar{A}_{i_0}^{y_0} > \bar{A}_{i_1}^1$ can only stem from Step 4 because $\bar{A}_{i_1}^1$ has the prevail condition of $\bar{A}_{i_0}^{y_0}$ as its effect, and $A_{i_0}^{x_0} < A_{i_1}^1$ can only stem from Step 6 because the prevail condition of $A_{i_0}^{x_0}$ is the precondition of $A_{i_1}^1$. From the relation between $A_{i_0}^{x_0}$ and $\bar{A}_{i_0}^{y_0}$, and the construction of $G'_e(v_{i_1})$ in FORWARD-CHECK, it follows that $\mathsf{post}(\bar{A}_{i_1}^1)$ appears before $\mathsf{post}(A_{i_1}^1)$ on $\sigma(v_{i_1})$, and again, from the local monotonicity it follows that $\mathsf{post}(\bar{A}_{i_1}^{y_1})$ appears before $\mathsf{post}(A_{i_1}^{x_1})$ on $\sigma(v_{i_1})$.

Alternatively, if $x_0 = y_0 = 1$ then $A_{i_0}^{x_0} = \bar{A}_{i_0}^{y_0} = A_i$. From (a) immediately follows that $A_{i_1}^1 \neq \bar{A}_{i_1}^1$, and an analysis similar to the above shows that $\mathsf{post}(\bar{A}_{i_1}^{y_1})$ appears before $\mathsf{post}(A_{i_1}^{x_1})$ on $\sigma(v_{i_1})$.

Having established that $\mathsf{post}(\bar{A}_{i_1}^{y_1})$ appears before $\mathsf{post}(A_{i_1}^{x_1})$ on $\sigma(v_{i_1})$, it is apparent that an inductive argument will allow us to show that for all $k > 0$ we have that $\mathsf{post}(\bar{A}_{i_k}^{y_k})$ appears before $\mathsf{post}(A_{i_k}^{x_k})$ on $\sigma(v_{i_k})$. Note that in particular it means that the operators $A_{i_k}^{x_k}$ and $\bar{A}_{i_k}^{y_k}$ are different, and this contradicts our assumption that $A_{i_m}^{x_m} = \bar{A}_{i_m}^{y_m}$. $\quad\square$

**Theorem 2** *Plan existence for* strips *planning problems with unary operators and directed-path singly connected causal graph is* np-*complete.*

**Proof:** First we show the membership in np. Let $\mathsf{MinPlanSize}(\Pi)$ denote the size of the minimal plan for a problem instance $\Pi$. Using the $\mathsf{MaxReq}$ property of the state variables, the following upper bound for $\mathsf{MinPlanSize}(\Pi)$ is straightforward from the Lemma 1:

$$\mathsf{MinPlanSize}(\Pi) \leq \sum_{v \in \mathcal{V}} \mathsf{MaxReq}(v) \leq n^2 \tag{5}$$

Thus, if we guess a minimal solution for a given solvable problem, we can verify it in low polynomial time.

The proof of the hardness is by polynomial reduction from 3-sat to the corresponding propositional plan generation problem with a directed-path singly connected causal graph. 3-sat is the problem of finding a satisfying assignment for a propositional formula in conjunctive normal form in which each conjunct (clause) has at most three literals.

Let $\mathcal{F} = C_1 \wedge \ldots \wedge C_n$ be a propositional formula belonging to 3-sat, and let $X_1, \ldots, X_m$ be the variables used in $\mathcal{F}$. An equivalent propositional planning problem with a directed-path singly connected causal graph can be constructed as follows: The variable set is $\mathcal{V} = \{X_1, \bar{X}_1, \ldots, X_m, \bar{X}_m\} \cup \{C_1, \ldots, C_n\}$. The variables $X_i$ and $\bar{X}_i$ have no predecessors in the causal graph, thus $\mathsf{pred}(X_i) = \mathsf{pred}(\bar{X}_i) = \{\emptyset\}$. In turn, for $1 \leq i \leq n$, $\mathsf{pred}(C_i) = \{X_{i_1}, \bar{X}_{i_1}, X_{i_2}, \bar{X}_{i_2}, X_{i_3}, \bar{X}_{i_3}\}$, where $X_{i_1}, X_{i_2}$, and $X_{i_3}$ are the variables that participate in the $i$th clause of $\mathcal{F}$. Finally, *Init* and *Goal* consist of false and true assignments to all variables in $\mathcal{V}$, respectively.

Let every operator $A \in \Lambda$ be presented as a three-tuple $\langle \{\mathsf{pre}\}, \{\mathsf{post}\}, \{\mathsf{prv}\} \rangle$ of pre, post, and prevail conditions respectively. Then, the corresponding operator set $\Lambda$ is specified as follows:

$$\begin{aligned}
\Lambda_{X_i} &= \{ \quad \langle \{f\}, \{t\}, \{\} \rangle \quad \} \\
\Lambda_{\bar{X}_i} &= \{ \quad \langle \{f\}, \{t\}, \{\} \rangle \quad \} \\
\Lambda_{C_i} &= \{ \quad \langle \{f\}, \{t\}, \{\alpha_1^i\} \rangle, \ \langle \{f\}, \{t\}, \{\alpha_2^i\} \rangle, \ \langle \{f\}, \{t\}, \{\alpha_3^i\} \rangle \quad \}
\end{aligned}$$

where $\alpha_j^i$ $(1 \leq j \leq 3)$ corresponds to the truth assignment on the variable $X_{i_j}$ that satisfies the $i$th clause of $\mathcal{F}$. Let $C_i = (X_1 \vee \bar{X}_2 \vee X_8)$. Then $\alpha_1^i = \{X_1 = t, \bar{X}_1 = f\}$, $\alpha_2^i = \{X_2 = f, \bar{X}_2 = t\}$, and $\alpha_3^i = \{X_8 = t, \bar{X}_8 = f\}$.

To illustrate the proposed reduction consider the following example. Formula $\mathcal{F}$ consist of 3 clauses: $(x_1 \vee \bar{x_2} \vee x_3) \wedge (x_1 \vee \bar{x_2} \vee x_4) \wedge (x_2 \vee \bar{x_3} \vee \bar{x_4})$. The causal graph of the corresponding planning problem is as follows:

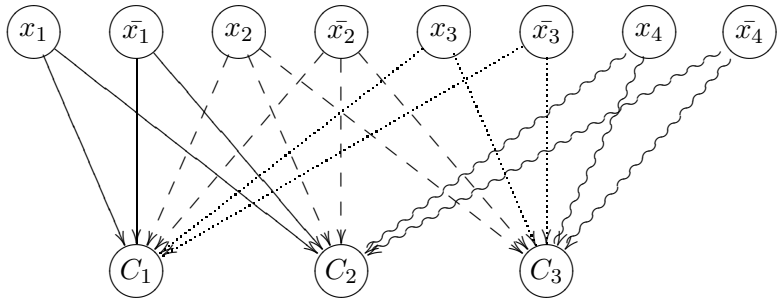

This is a propositional planning problem with single-effect operators and an underlying directed-path singly connected causal graph. Clearly, *Goal* is reachable ($\Pi$ is solvable) if and only if a satisfying assignment for $\mathcal{F}$ can be found. Thus, plan existence for the propositional planning problems with directed-path singly connected causal graphs is NP-complete. $\square$

**Lemma 4** *Given $k$ ordered sequences $\sigma_1, \cdots, \sigma_k$ of $n$ elements each, the number $T[k]$ of different merges of $\sigma_1, \cdots, \sigma_k$, preserving the orderings induced by $\sigma_1, \cdots, \sigma_k$ on their elements, is given by:*

$$T[k] = \prod_{i=1}^{k-1} \sum_{j=1}^{n} \binom{n-1}{j-1}\binom{ni+1}{j} \tag{6}$$

**Proof:** Considering the merge operation of such $k$ sequences as iterative merge of $\sigma_i$, $2 \leq i \leq k$, with the already merged sequences $\sigma_1, \ldots, \sigma_{i-1}$, it is easy to see that $T(k)$ can be expressed as:

$$T[k] = \begin{cases} T[k-1] \cdot S\left[n(k-1), n\right], & k > 1 \\ 1, & k = 1 \end{cases} \tag{7}$$

where $S[x, y]$ stands for the number of different, order preserving merges of *two* ordered sequences of sizes $x$ and $y$ (without loss of generality, we assume that $x \geq y$).

We consider the process of merging two ordered sequences $\sigma$ and $\sigma'$, $|\sigma| \geq |\sigma'|$, as:

(i) partition of $\sigma'$ into $j$ sub-sequences,

(ii) partition of $\sigma$ into $l$ sub-sequences, where $j - 1 \leq l \leq j + 1$, and

(iii) *interleaving* and order preserving concatenation of the sub-sequences of $\sigma$ and $\sigma'$.

First, observe that $\sigma'$ can be partitioned into $1 \leq j \leq |\sigma'|$ sub-sequences. Second, for any $j$, the numbers of different partitions corresponding to steps (i) and (ii) are $\binom{|\sigma'|-1}{j-1}$ and $\binom{|\sigma|+1}{j}$, respectively. Finally, given a pair of such partitions of $\sigma$ and $\sigma'$, there exist exactly one possible interleaving and order preserving concatenation as in step (iii). Therefore, we have:

$$S(x, y) = \sum_{j=1}^{y} \binom{y-1}{j-1}\binom{x+1}{j} \tag{8}$$

and combining Eq. 7 with Eq. 8, we arrive to Eq. 6. $\square$